\pdfoutput=1

\documentclass[10pt,twocolumn,letterpaper]{article}
\usepackage[pagenumbers]{cvpr} % To force page numbers, e.g. for an arXiv version

\usepackage{times}
\usepackage{epsfig}
\usepackage{graphicx}
\usepackage{amsmath}
\usepackage{amssymb}
\usepackage{booktabs}

\usepackage{csquotes}
\usepackage[dvipsnames]{xcolor}
\usepackage{caption}
\usepackage{xparse}
\usepackage{xfrac}
\usepackage{bm}
\usepackage[normalem]{ulem} 
\usepackage[flushleft]{threeparttable}
\usepackage{gensymb}
\usepackage{wrapfig}
\usepackage{dsfont}

\definecolor{cvprblue}{rgb}{0.21,0.49,0.74}
\usepackage[pagebackref,breaklinks,colorlinks,citecolor=cvprblue]{hyperref}
\usepackage[capitalize]{cleveref}

\newtoggle{nographics}
\newtoggle{lrgraphics}
\newtoggle{nocomments}

\togglefalse{nographics}
\togglefalse{lrgraphics}
\togglefalse{nocomments}
\crefname{section}{Section}{Sections}
\crefname{table}{Table}{Tables}
\crefname{figure}{Figure}{Figures}
\Crefname{section}{Section}{Sections}

\setlist{topsep=0pt,itemsep=-5pt,leftmargin=*}
\begin{document}

%%%%%%%%%%%%%%%%%%%%%%%%%%%%%%%%%%%%%%%%%%%%%%%%%%%
%%%   P macros %%%%%%%%%%%%%%%%%%%%%%%%%%%%%%%%%%%%
%%%%%%%%%%%%%%%%%%%%%%%%%%%%%%%%%%%%%%%%%%%%%%%%%%%
\newcommand{\Kill}[1]{}
%! Suppress = NonMatchingIf
\iftoggle{nocomments}{
    \newcommand{\NS}[1]{\ignorespaces}
    \newcommand{\DZ}[1]{\ignorespaces}
    \newcommand{\MI}[1]{\ignorespaces}
    \newcommand{\LA}[1]{\ignorespaces}
    \newcommand{\AI}[1]{\ignorespaces}
    \newcommand{\AK}[1]{\ignorespaces}
    \newcommand{\EB}[1]{\ignorespaces}
    \newcommand{\todo}[1]{\ignorespaces}
    \newcommand{\plan}[1]{\ignorespaces}
}{
    \newcommand{\NS}[1]{\textbf{\textcolor{violet}{NS: #1}}}
    \newcommand{\DZ}[1]{\textbf{\textcolor{purple}{DZ: #1}}}
    \newcommand{\MI}[1]{\textbf{\textcolor{cyan}{MI: #1}}}
    \newcommand{\LA}[1]{\textbf{\textcolor{orange}{AA: #1}}}
    \newcommand{\AK}[1]{\textbf{\textcolor{green}{AK: #1}}}
    \newcommand{\AI}[1]{\textbf{\textcolor{brown}{AI: #1}}}
    \newcommand{\EB}[1]{\textbf{\textcolor{blue}{EB: #1}}}
    \newcommand{\todo}[1]{\textbf{\textcolor{red}{#1}}}
    \newcommand{\plan}[1]{\textsl{\textcolor{gray}{Say: #1}}}
}

%%%%%%%%%%%%%%%%%%%%%%%%%%%%%%%%%%%%%%%%%%%%%%%%%%%
%   Graphics %%%%%%%%%%%%%%%%%%%%%%%%%%%%%%%%%%%%%%
%%%%%%%%%%%%%%%%%%%%%%%%%%%%%%%%%%%%%%%%%%%%%%%%%%%
\DeclareGraphicsExtensions{.eps,.pdf,.jpg,.png}
\iftoggle{lrgraphics}{
    \graphicspath{{src/media/lr/}{src/media/vector/}}
}{
    \graphicspath{{src/media/hr/}{src/media/vector/}}
}
\makeatletter
\iftoggle{nographics}{
    \LetLtxMacro{\includegraphics@orig}{\includegraphics}
    \RenewDocumentCommand{\includegraphics}{ s O{} m }{%
        {\setlength{\fboxsep}{0pt}%
         \colorbox{lightgray}{\phantom{\IfBooleanTF{#1}{\includegraphics@orig*}{\   includegraphics@orig}[#2]{#3}}}%
        }%
    }
}{}
\makeatother

%%%%%%%%%%%%%%%%%%%%%%%%%%%%%%%%%%%%%%%%%%%%%%%%%%%
%   Math symbols %%%%%%%%%%%%%%%%%%%%%%%%%%%%%%%%%%
%%%%%%%%%%%%%%%%%%%%%%%%%%%%%%%%%%%%%%%%%%%%%%%%%%%
\NewDocumentCommand\bv{}{\bm{v}}
\NewDocumentCommand\bu{}{\bm{u}}
\NewDocumentCommand\bp{}{\bm{p}}
\NewDocumentCommand\br{}{\bm{r}}
\NewDocumentCommand\bn{}{\bm{n}}

%%%%%%%%%%%%%%%%%%%%%%%%%%%%%%%%%%%%%%%%%%%%%%%%%%%
%   Quality measures %%%%%%%%%%%%%%%%%%%%%%%%%%%%%%%%%%
%%%%%%%%%%%%%%%%%%%%%%%%%%%%%%%%%%%%%%%%%%%%%%%%%%%
\NewDocumentCommand\rmse{}{%
    \ifmmode \text{RMSE}
    \else RMSE\xspace 
    \fi %
}
\NewDocumentCommand\qrmse{}{%
    \ifmmode \text{RMSE-}q_{95} 
    \else RMSE-$q_{95}$\xspace 
    \fi %
}
\NewDocumentCommand\fpr{}{%
    \ifmmode \text{FPR} 
    \else FPR\xspace 
    \fi %
}
\NewDocumentCommand\recall{}{%
    \ifmmode \text{Recall} 
    \else Recall\xspace 
    \fi %
}
\NewDocumentCommand\diracc{}{%
    \ifmmode \text{AngleAcc}
    \else AngleAcc\xspace 
    \fi %
}

\NewDocumentCommand\rmsev{}{%
    \ifmmode \text{RMSE}_{\text{v}}$
    \else RMSE$_{\text{v}}$\xspace 
    \fi %
}
\NewDocumentCommand\lpipsv{}{%
    \ifmmode \text{LPIPS}_{\text{v}}
    \else LPIPS$_{\text{v}}$\xspace 
    \fi %
}

\NewDocumentCommand\samplingdistance{}{%
    \ifmmode \lambda
    \else $\lambda$\xspace 
    \fi %
}

\title{PRS: Sharp Feature Priors for Resolution-Free Surface Remeshing}

\author{Natalia Soboleva\textsuperscript{1}~~~
Olga Gorbunova\textsuperscript{1,3}~~~
Maria Ivanova\textsuperscript{1,2}~~~
Evgeny Burnaev\textsuperscript{1,2}\\
Matthias Nie{\ss}ner\textsuperscript{4}~~~
Denis Zorin\textsuperscript{5}~~~
Alexey Artemov\textsuperscript{4}
\vspace{.2em} \\
\textsuperscript{1}Skolkovo Institute of Science and Technology~~~
\textsuperscript{2}Artificial Intelligence Research Institute\\
\textsuperscript{3}Yandex LLC~~~
\textsuperscript{4}Technical University of Munich~~~
\textsuperscript{5}New York University\\
}

\twocolumn[{%
    \renewcommand\twocolumn[1][]{#1}%
    \vspace{-40pt}
    \maketitle
    \vspace{-30pt}
    \begin{center}
        \includegraphics[width=\linewidth]{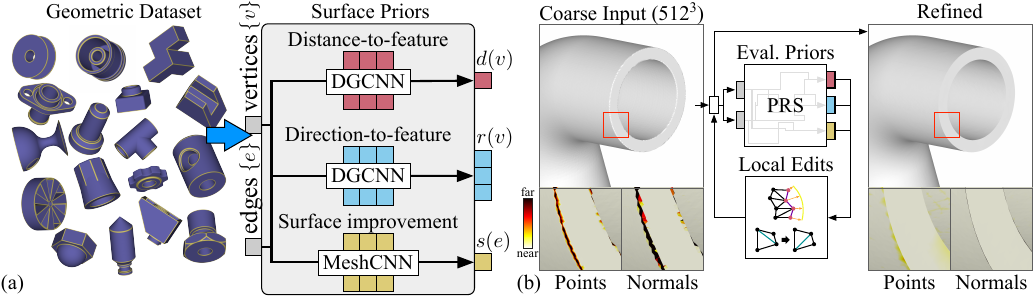}
        \captionof{figure}{We use large volumes of geometric data to learn a collection of neural models sensitive to reconstruction accuracy of sharp geometric features~(a), that we view as surface-based priors (PRS). 
        We integrate our priors in an unstructured mesh editing algorithm for feature-aware remeshing of raw aliased mesh reconstructions, resulting in a faithful approximation of sharp geometric features (b).}
        \label{fig:teaser}
    \end{center}    
}]

%%%%%%%%% ABSTRACT
\begin{abstract}
Surface reconstruction with preservation of geometric features is a challenging computer vision task. 
Despite significant progress in implicit shape reconstruction, state-of-the-art mesh extraction methods often produce aliased, perceptually distorted surfaces and lack scalability to high-resolution 3D shapes. 
We present a data-driven approach\footnote{\href{https://artonson.github.io/publications/2024-prs/}{artonson.github.io/publications/2024-prs}} for automatic feature detection and remeshing that requires only a coarse, aliased mesh as input and scales to arbitrary resolution reconstructions. 
We define and learn a collection of surface-based fields to (1) capture sharp geometric features in the shape with an implicit vertexwise model and (2) approximate improvements in normals alignment obtained by applying edge-flips with an edgewise model. 
To support scaling to arbitrary complexity shapes, we learn our fields using local triangulated patches, fusing estimates on complete surface meshes.
Our feature remeshing algorithm integrates the learned fields as sharp feature priors and optimizes vertex placement and mesh connectivity for maximum expected surface improvement. 
On a challenging collection of high-resolution shape reconstructions in the ABC dataset~\cite{koch2019abc}, our algorithm improves over state-of-the-art by 26\% normals F-score and 42\% perceptual \rmsev. 
\end{abstract}

\section{Introduction}
\label{sec:intro}

Accurate 3D surface mesh models are commonly viewed as a desired final representation for 3D reconstruction; this view is supported by the prevalence of algorithmic and software tools that facilitate mesh manipulation. 
An important requirement for precise, high-quality surface reconstruction is preserving and explicitly representing sharp geometric features: perceptually, structurally, and functionally crucial regions
present in many man-made shapes and explicitly stored in the CAD models~\cite{koch2019abc,matveev2022def}
but lost in raw meshes generated by 3D scanning devices or by iso-surfacing~\cite{lorensen1987marching,lewiner2003efficient,ju2002dual} volumetric representations.

Surface reconstruction with preservation of sharp geometric features has been approached from multiple directions, with notable improvements achieved in feature estimation~\cite{himeur2021pcednet,matveev2022def,wang2020pie}, surface reconstruction~\cite{xu2022rfeps,feng2023deep} and meshing~\cite{chen2021neural,chen2022neural}, or feature-preserving mesh smoothing~\cite{hattori2022learning,shen2022gcn,wei2018mesh,zhang2015guided} and re-meshing~\cite{jakob2015instant}.
However, most approaches come with practical limitations: they require access to additional information (\eg, clean 3D scans or accurate implicit functions), need manual, per-instance parameter tuning, or assume features to be already available (\eg from prior estimation or user input). 
Many recent methods target iso-surfacing of implicit functions sampled on medium-size $128^3$ grids and do not support high-resolution grids. As a result, despite substantial effort, reconstructing feature-equipped meshes remains a challenging task.

One promising path forward is to start with an initial, coarse mesh approximation constructed by an established procedure (\eg,~\cite{lorensen1987marching,lewiner2003efficient,ju2002dual}) and to iteratively refine it by applying standard local \textit{remeshing} operations such as flipping edges and relocating vertices~\cite{alliez2008recent,surazhsky2003explicit} (for the example in the inset, performing six vertex moves and two edge-flips 
perfectly aligns the edges %in the input mesh 
\begin{wrapfigure}{r}{0.6\columnwidth}
  \vspace{-10pt}
  \begin{center}
    \includegraphics[
    width=0.6\columnwidth,
    trim=2pt 2pt 2pt 2pt, clip=True,
    ]{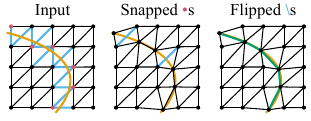}
  \end{center}
  \vspace{-20pt}
\end{wrapfigure}
to a given curve). 
However, the detection of feature regions, selection of individual elements, and identification of parameters for each edit operation (\eg, deciding where to offset each vertex) all represent important challenges.  

We argue that learning-based methods are a natural fit for such tasks as they can effectively leverage statistical patterns in large mesh datasets to predict improvements expected from applying editing operations in each location, thus serving as a form of surface priors for remeshing. 
In addition, they have the advantage of automatically adapting to individual input shapes without requiring per-model region-of-interest specification, supporting scaling to complex, high-resolution 3D models. 
Recent deep learning architectures operate directly on meshes or proxy graphs~\cite{hanocka2019meshcnn,wang2019dynamic}, avoiding the need to involve any data other that the mesh itself.

In this paper, we propose an algorithm for refinement of triangle meshes designed to 
(1) automatically detect and reconstruct sharp geometric features in meshes, 
(2) require no user interaction for processing a particular input shape, 
(3) require no inputs other than triangle meshes, 
and (4) scale to 3D shapes meshed at arbitrary resolution. 
To satisfy these desiderata, we make a number of design decisions that we summarize below.

Our feature detection algorithm uses three fields defined at the elements of the input (coarse) surface mesh: two vertex-based fields encoding distances and normalized directions to the closest points on the feature curves, and an edge-based field capturing improvements in normal alignment obtained by flipping each edge, provided that all other remain fixed. 
We train three mesh-based neural networks to approximate each field.
To scale our approach to complex 3D shapes, we learn all three fields on local triangulated patches; for reconstructing fields on a complete surface mesh, we fuse estimates from overlapping patches. 
We train our networks using a large-scale dataset of level-set meshes with realistic irregular chamfering and connectivity artifacts obtained from a simulated reconstruction process.
For this purpose, we develop a training data generation pipeline based on high-quality, feature-equipped surface meshes in the ABC dataset~\cite{koch2019abc}.

Building on the three fields defined on complete 3D shapes, we develop an iterative algorithm integrating the fields as priors to achieve accurate feature reconstruction and remeshing. 
Specifically, in contrast to most mesh denoising methods, our remeshing algorithm optimizes \textit{both vertex placement and edge connectivity} by snapping closest vertices to features and flipping edges with largest expected surface improvements. 
At each iteration, the procedure alternates evaluation of trained networks with local remeshing, checks predicted quantities against a threshold, and stops when expected gains in surface quality are small. %drop to small values. 

We benchmark our algorithm against state-of-the-art mesh extraction methods~\cite{kobbelt2001feature,chen2021neural,chen2022neural} and combinations of standard mesh extraction with feature-preserving denoising~\cite{zhang2015guided,shen2022gcn}, focusing on fine-grained feature reconstruction and surface quality measures. 
We additionally demonstrate how our algorithm complements smoothing methods, serving as a processing step in a reconstruction pipeline.

We contribute:
\begin{itemize} 
\item An automatic method for estimating sharp geometric features and identifying surface meshing artifacts in 3D surface mesh models of arbitrary resolution, utilizing localized predictions and globally fusing local estimates.

\item An automatic method for reconstructing and remeshing sharp geometric features in 3D surface meshes incorporating feature and surface improvement fields as priors for local surface remeshing.
\end{itemize}

\section{Related Work}
\label{sec:related}

\noindent \textbf{Surface Reconstruction} is a central task in geometry processing~\cite{berger2014state,khatamian2016survey,huang2022surface}. 
Generating surfaces \textit{explicitly,} \eg, by Delaunay triangulation of raw point sets~\cite{cazals2006delaunay,digne2011scale} or using recent neural network (NN)-based approaches~\cite{sharp2020pointtrinet,liu2020meshing,rakotosaona2021learning} has limited use as these methods do not produce watertight, manifold structures needed by many applications.

\textit{Implicit} reconstruction involves (1) fitting a scalar field to input points and (2) extracting its iso-surface as a mesh. 
A variety of methods including moving least squares regression~\cite{cheng2008survey,oztireli2009feature}, Poisson surface reconstruction (PSR)~\cite{kazhdan2013screened}, and deep kernel regression~\cite{williams2022neural} are available for approximating scalar fields, each with trade-offs between robustness and detail preservation.
Neural implicit fields, using NNs to represent the scalar field globally~\cite{park2019deepsdf,mescheder2019occupancy}, or locally~\cite{jiang2020local,chabra2020deep}, improve fitting quality and scalability of implicit representations in many instances.
For mesh extraction, all these approaches rely on the second ingredient of implicit reconstruction, iso-surfacing algorithms.

Iso-surfacing refers to approximating level sets of implicit functions (\eg, a signed distance field (SDF)).
Traditionally, SDF values are sampled and analyzed in few individual, isolated locations such as vertices of cubes~\cite{lorensen1987marching} or tetrahedra~\cite{gueziec1995exploiting,treece1999regularised}.  Marching cubes (MC33)~\cite{lorensen1987marching,chernyaev1995marching,lewiner2003efficient} enumerates sign configurations of the implicit field in vertices of a cube, defining the surface passing though points  trilinearly interpolated on its edges. 
Such algorithms are local and straightforward to implement but lack guarantees on meeting feature preservation or mesh quality conditions. 
Numerous improvements have been proposed in recent decades to address these limitations~\cite{chen2021neural,newman2006survey,de2015survey}.

Relevant to this work, for preservation of sharp geometric features, several methods integrate alternative implicit representations (\eg, distance gradients in addition to SDF values~\cite{ju2002dual,kobbelt2001feature}) or adaptive grid generation~\cite{qu2004feature,varadhan2003feature}. 
These methods require explicitly defining features to generate an initial mesh. 
We experimentally compare to EMC~\cite{kobbelt2001feature}.

Recently, Neural Marching Cubes (NMC)~\cite{chen2021neural} and Neural Dual Contouring (NDC)~\cite{chen2022neural} enhanced the triangulation algorithm with broader contextual information by training on a dataset of meshes.
% While improving surface extraction quality compared to the MC33 baseline, 
These methods lack scalability to 3D shapes sampled beyond the resolution they were trained for ($128^3$). 
In contrast, our algorithm naturally processes meshes sampled at arbitrary resolution. 
We experimentally show that our feature-aware algorithm results in more accurate surface reconstruction, especially near features.

\begin{figure*}[t]
\centerline{\includegraphics[
width=.925\textwidth,
trim=0 0.5em 0 0, 
clip=True]{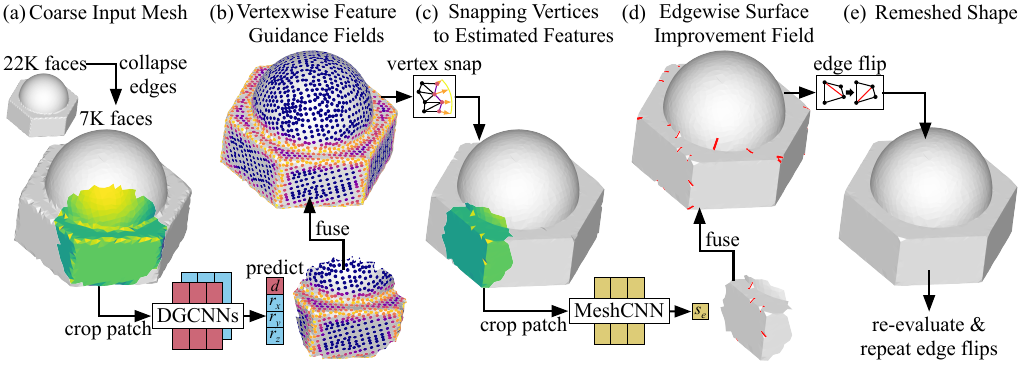}}
\caption{We start with an coarse input mesh that we decimate by collapsing short edges (a). 
To detect poorly meshed near-feature regions, we crop local triangulated mesh patches, predict patch-level distance- and direction-to-feature fields~\eqref{eq:feature_guidance} 
and integrate these into an object-level field (b); 
we use this field to snap near-feature vertices onto predicted feature curves (c).
To further improve feature meshing, we iteratively detect edges with large potential surface improvement~\eqref{eq:surface_improvement} and flip these (d--e). 
Mesh integrity is maintained at each stage.
}
\label{fig:method-overview}
\end{figure*}

\noindent \textbf{Mesh Denoising} focuses on removing noise from vertex positions while retaining geometric features~\cite{attene2013polygon,chen2022geometric,sun2007fast}. 
Many works adopt a two-step process, including filtering facet normals, often using a bilateral filter~\cite{zheng2010bilateral}, and updating vertices to match denoised normals~\cite{taubin2001linear}.
The effectiveness of the bilateral filter relies on the range kernel; constructing guidance normals was proposed through fitting quadrics~\cite{fan2009robust,wang2012cascaded}, bilateral filtering~\cite{zheng2010bilateral,yadav2018robust}, patch-based normal filtering~\cite{zhang2015guided}, or non-local filtering~\cite{li2018non,wei2018mesh}.
Recent data-driven mesh denoising methods learn guidance normals from training shapes~\cite{wang2016mesh,zhao2019normalnet,shen2022gcn,armando2020mesh,zhang2022geobi}. 

Denoising per-vertex noise presumes that sharp features are present but corrupted by noise; this task is different from ours where features are irregularly triangulated and require reconstruction. 
As many denoising methods distort features when used in isolation~\cite{chen2022geometric}, we demonstrate how one can complement these approaches using our feature-aware algorithm, integrating it with a two-stage denoising filter~\cite{zhang2015guided}.

\noindent \textbf{Feature Detection. }
Commonly, sets of mesh vertices are identified as instances of specific features (\eg, sharp features, ridges, valleys, \etc).
Estimating such features involves computing derivatives of principal curvature, a highly unstable computation for discrete, possibly noisy, surfaces~\cite{hildebrandt2005smooth,kalogerakis2007robust,lai2006robust,yoshizawa2008fast}. 
Methods with robustness to sampling and noise exist~\cite{ohtake2004ridge,kim2009feature,torrente2018recognition}, but rely on proper feature reconstruction during mesh generation. 
Recent learning-based feature detection approaches operate directly on raw 3D scans~\cite{himeur2021pcednet,liupc2wf,wang2020pie,matveev2022def}; however, incorporating feature estimates in surface reconstructions has only seen few initial attempts~\cite{feng2023deep,xu2022rfeps}. 
Our method reconstructs sharp features directly on meshes but does not expect features to form sequences of vertices~\cite{thompson2019shrec}, be directly sampled, or even belong to the surface.
Instead, we view features as curves in 3D space, generally extending outside the model, and adopt an implicit feature representation similar to~\cite{matveev2022def}.

\noindent \textbf{Feature Remeshing} aims to improve the approximation of surface features in the mesh. 
Methods that extrapolate feature-adjacent smooth regions~\cite{attene2003edge} or snap vertices to detected features~\cite{vorsatz2001feature,tarini2010practical} were proposed. 
Recently, feature preservation has been shown via smooth, global mesh-field alignment~\cite{jakob2015instant,pietroni2021reliable}. 
All these approaches require that features are supplied as input in some form. 
In contrast, our framework integrates automatic feature detection and remeshing in a unified algorithm.

\section{Learning Sharp Feature Priors}
\label{sec:priors}

\subsection{Field Representations and Data Generation}
\label{sec:datasets}

We represent triangular meshes by a triple $(V, F,E)$, where  $V = \{ v_i \}$ is the set of vertices, specified by their 3D coordinates, and $F = \{ f_j \}$  is the set of faces given by triples of indexes into $V$, and the set of edges $E = \{ e_k \}$, given by vertex pairs.

\noindent \textbf{Feature-Equipped 3D Shapes.}
Our source data is a subset of high-quality, feature-aligned 3D meshes provided in the ABC dataset~\cite{koch2019abc}, constructed by sampling and triangulating parametric boundary representations (B-Reps) of the source CAD shapes, and providing vertexwise sharp feature annotations obtained by thresholding differences in analytic normals in smooth adjacent CAD regions in the B-Rep.

\noindent \textbf{Construction of Coarse 3D Meshes.}
Per-vertex noise~\cite{li2020dnf,shen2022gcn} or random edge flips~\cite{hanocka2019meshcnn} do not model well surface reconstruction flaws such as irregular chamfering of sharp features~\cite{alliez2008recent}. 
Instead, we simulate a simplified reconstruction process for obtaining realistic mesh extraction artifacts. 
We build a volumetric SDF with voxel size~$\samplingdistance$ by querying point-to-surface distances from each grid point of a regular 3D voxel grid to the ground-truth (GT) mesh and extract its level set using MC33~\cite{lorensen1987marching,lewiner2003efficient} as the coarse mesh.
To model unknown orientation, each GT mesh is randomly rotated before computing the SDF. 
Unlike methods assuming a fixed grid resolution (\eg, $64^3$~\cite{chen2021neural,chen2022neural}), we adapt the resolution of the volumetric grid to the complexity of the GT mesh, scaling it to sample feature curves of characteristic linear size with at least $n$ points.
We prune narrow triangles and remove excess vertices by collapsing short edges of length $\alpha_{\text{short}} \samplingdistance$ or smaller.
Memory requirements are bounded by pre-selecting a set of models in ABC that yield volumetric grids of up to $512^3$ voxels.
We view $512^3$ as coarse despite high voxel resolution due to aliasing artefacts (see explanation in~\cref{fig:resolution_dependent_measures}).

\noindent \textbf{Mesh Correspondence Maps.}
To compute edge- and vertex-based fields, we transfer feature- and surface-aware annotations to the coarse mesh by constructing two correspondence maps between the faces of the GT and the newly constructed coarse surfaces. 
The coarse-to-GT map ($\pi_{\text{c}\to\text{g}}$) is computed by uniformly sampling the coarse mesh and finding faces on the GT mesh closest to each sample. 
Swapping the two meshes yields the GT-to-coarse map ($\pi_{\text{g}\to\text{c}}$). 
Bidirectional mapping accurately handles instances of differences in face sizes between the two meshes.
% Importantly, mappings enable the transfer of feature- and surface-aware annotations to the coarse mesh by integrating over reference elements mapped to by $\pi_{\text{c}\to\text{r}}$.

\noindent \textbf{Patch and Feature Selection.}
As training with detailed 3D meshes is impractical and is likely to generalize poorly, we train on a set of local mesh patches, constraining the complexity of training instances and enabling handling of arbitrarily large models. 
For a coarse surface, we sample up to 1K uniformly distributed seed points, cut out connected sub-meshes within the distance $\alpha_{\text{patch}} \samplingdistance$ to each seed, and retrieve their feature-equipped counterparts from the GT mesh using $\pi_{\text{c}\to\text{g}}$.
We keep interior sharp features (\ie, those passing through non-boundary edges in the GT patch) and eliminate the rest.

\noindent \textbf{Distance- and Direction-to-Feature Fields.}
Following a recent feature estimation approach~\cite{matveev2022def}, we implicitly represent features using the scalar \emph{distance-to-feature} field $d^{\varepsilon}(v)$ and the vector \emph{direction-to-feature} field $r^{\varepsilon}(v)$, defining these at vertices $v \in V$ in the coarse mesh. 
We truncate both fields at a radius~$\varepsilon$, a multiple of the sampling distance~$\lambda$ (choosing $\varepsilon = 4 \samplingdistance$).
To compute these quantities at a vertex $v$, we map $v$ to its closest neighbor $q(v)$ belonging to the feature curves in the GT mesh and set 
\begin{align}
\label{eq:feature_guidance}
d^{\varepsilon}(v) &= \min\big( ||q(v) - v||, \varepsilon \big) \nonumber \\
r^{\varepsilon}(v) &= \mathrm{normalize}(q(v) - v) \cdot \mathbb{I}(d^{\varepsilon}(v) < \varepsilon).
\end{align}
where $\mathbb{I}$ denotes the indicator function.
The field~$r^{\varepsilon}$ contains unit-length direction vectors for points closer to features than~$\varepsilon$, and zeros elsewhere.

\noindent \textbf{Surface Improvement Field} captures changes in normals alignment between the coarse and the GT meshes resulting from an isolated edge-flip. 
We start by defining a scalar per-face normal consistency (NC) measure by integrating deviation between the coarse normal~$n_{\text{crs}}(f)$ and GT normals sampled in matching (under $\pi_{\text{c}\to\text{g}}$) locations:
\begin{equation}
\label{eq:face_normals_consistency}
\mathrm{nc}(f) = \frac{1}{|f|} \int_f 
    \left\lVert
    n_{\text{crs}}(u) - 
      n_{\text{gt}}(\pi_{\text{c}\to\text{g}}(u)) 
    \right\rVert 
     du.
\end{equation}
Edge-based normals consistency $\mathrm{nc}(e)$ is given by weighted averaging $\mathrm{nc}(e) = w_1 \mathrm{nc}(f_1) + w_2 \mathrm{nc}(f_2)$ of the two face-based values $\mathrm{nc}(f_{1,2})$ incident at~$e$. 
Finally, our edgewise surface improvement field is given by the difference
\begin{equation}
\label{eq:surface_improvement}
s(e) = \mathrm{nc}(e') - \mathrm{nc}(e), \quad e \in E,
\end{equation}
where~$\mathrm{nc}(e')$ refers to the NC of the flipped edge~$e'$.

\subsection{Patch-Based Field Estimation and Fusion}
\label{sec:learning}

We learn signals directly on triangular meshes, without resorting to regular representations (\eg, voxelization or sampling). 
We introduce three mesh-based deep networks $f_{\text{d}}$, $f_{\text{r}}$, and $f_{\text{s}}$ for regressing distance-to-feature, direction-to-feature, and surface improvement fields, respectively.

\noindent \textbf{Distance- and Direction-to-Feature Fields.}
We define our deep estimators of distance- and direction-to-feature fields at vertices~$V$ in the coarse mesh. 
Architecturally, our models are generalized variants of the DGCNN~\cite{wang2019dynamic}. 
We find the optimal learning configuration by experimenting with hyperparameters summarized in~\cref{tab:ablation-dists-dirs}. 
The distance-to-feature model optimizes a combination of $L_2$ and Laplacian terms emphasizing accurate, smooth predictions for vertices closest to the sharp features: 
\begin{equation}
\label{eq:distances_loss}
L(d^{\varepsilon}, \widehat{d}^{\varepsilon}) 
= L_2 (\log d^{\varepsilon}, \log \widehat{d}^{\varepsilon}) 
+ 0.1 L_{\text{lap}}(d^{\varepsilon}, \widehat{d}^{\varepsilon}), 
\end{equation}
In this equation, the Laplacian smoothing term is
\begin{equation}
\label{eq:lap_loss}
L_{\text{lap}}(d^{\varepsilon}, \widehat{d}^{\varepsilon}) = 
(\Delta d^{\varepsilon} - \Delta \widehat{d}^{\varepsilon})^2,
\end{equation}
where the Laplacian of $d^{\varepsilon}$ can be computed by
\begin{equation}
\label{eq:laplacian}
(\Delta d^{\varepsilon})(v) = \frac{1}{Z}\sum_{u \in N(v)} L_{vu} d^{\varepsilon}(u) - d^{\varepsilon}(v). 
\end{equation}
Here $Z = \sum_{u \in N(v)} L_{vu}$ is the normalizing constant, $N(v)$ is the set of vertices adjacent to $v$, and $L_{vu}$ is the discrete cotangent Laplacian (see, \eg,~\cite{crane2013digital}).
Our directions-to-feature estimator minimizes $L_2$ loss for vertices.

\noindent \textbf{Surface Improvement Field.}
We define our deep surface improvement estimator at the edges~$E$ of the coarse patches and uses these predictions for edge flips during remeshing. 
We use the edge-based MeshCNN network~\cite{hanocka2019meshcnn}, modifying it to perform fewer poolings (1.2K vertices) to avoid collapsing boundary edges. 
We train our network using the binary cross-entropy loss to predict surface improvement values~$s(e)$ exceeding the threshold $\alpha_s = 0.1$.

\begin{table}[t]
\centering
\includegraphics[width=\columnwidth]{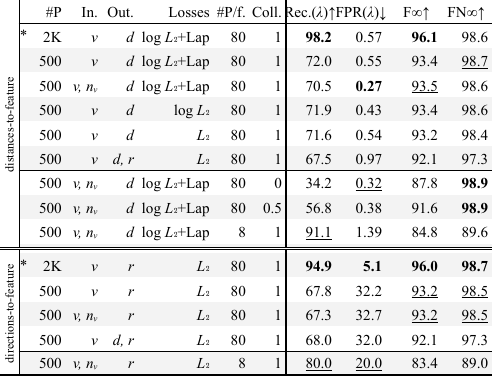} 
\caption{Quantitative comparison of various learning configurations for estimating distance \textit{(top)} and direction \textit{(bottom)} fields. 
\#P: number of vertices, 
Inputs (In.): vertices ($v$), vertex normals $n_v$, 
Outputs (Out.): distances ($d$), directions ($r$), 
\#P/f.: points per feature, 
Collapses (Coll.): ratio of collapsed edges (\cref{sec:datasets}).}
\label{tab:ablation-dists-dirs}
\end{table}

\noindent \textbf{Fusion of Per-Patch Predictions.}
We reconstruct the three fields $\widehat{d}, \widehat{r}, \widehat{s}$ on complete 3D meshes from local patch-based predictions.
Patches outside feature regions are excluded if $q_{\alpha}(\widehat{d}_i) > \samplingdistance /2$ where $q_{\alpha}(\widehat{d}_i)$ represents the $\alpha$-quantile of values in $\widehat{d}_i$ (we use $\alpha = 10\%$).
Multiple predictions in each vertex are correlated due to overlapping patches. We use a statistical algorithm for averaging correlated predictions~\cite{schmelling1995averaging}, involving computing weighted averages of field values in each element in the mesh.

\section{Mesh Refinement}
\label{sec:remeshing}

\noindent \textbf{Overview.}
Our remeshing method iteratively optimizes vertex locations and face orientations to reconstruct cleaner sharp geometric features.
We use standard remeshing operations, \emph{vertex repositioning} and \emph{edge flips} (see~\cite{hoppe1993mesh,botsch2010polygon}). 
The application of these operations is guided by fields predicted by our neural networks. Additionally, we incorporate \emph{edge and face collapses}~\cite{hoppe1996progressive} as preliminary steps to simplify input meshes and match the training data distribution.
To mitigate undesired effects during the remeshing process~-- such as abrupt flips in face normals or  manifoldness violations-- we establish and maintain a set of invariants. 
These invariants are checked for each modified element, following a simplified scheme inspired by~\cite{jiang2022declarative}. 
Any edits violating these invariants are discarded in each iteration before applying them to the input mesh.
\cref{fig:method-overview} presents an overview of steps in our remeshing procedure, described below.

\noindent \textbf{Patch Extraction.}
We densely extract local, overlapping triangulated patches ${P_i}$ of up to $n_v = 2000$ vertices from the input mesh. 
Local feature estimation allows our method to scale effectively to arbitrarily large meshes. 
The overlap across patches serves as a form of test-time augmentation~\cite{shorten2019survey}, enhancing the robustness of our approach.

\noindent \textbf{Patch-Based Feature Estimation and Fusion.}
We use the trained networks~$f_{\text{d}}, f_{\text{r}}$ to predict a scalar distance-to-feature field~$\widehat{d}^{\varepsilon}_i$ and a vector direction-to-feature field~$\widehat{r}^{\varepsilon}_i$ at vertices $V = \{v\}$ in each patch~$P_i$.
We fuse per-patch predictions to reconstruct the shape-level feature-aware fields.

\begin{figure}[t!]
\centerline{\includegraphics[
width=\columnwidth,
trim=0 0 0 1.5em,
clip=True]{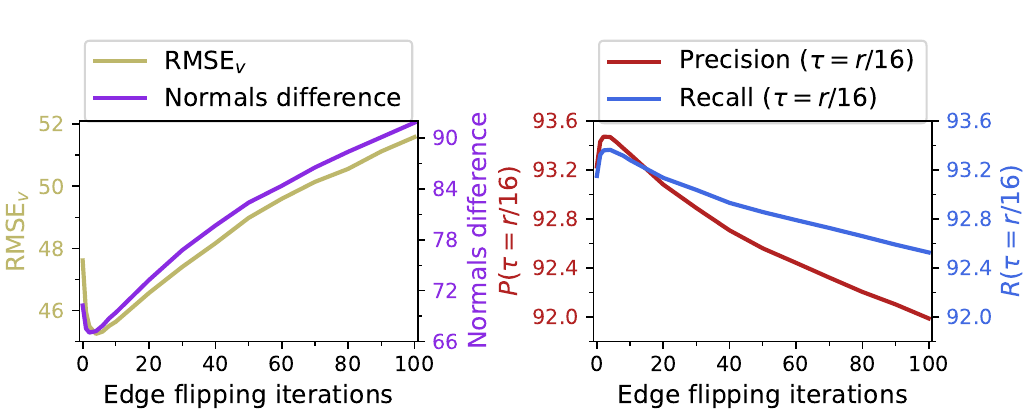}}
\caption{Empirically, only a few (2--4) sets of edge flips yield improvements across most measures as meshes during this stage most closely correspond those seen in training set.}
\label{fig:quality-edge-flipping-iterations}
\end{figure}

\begin{table*}[t]
\centering
\includegraphics[width=\textwidth]{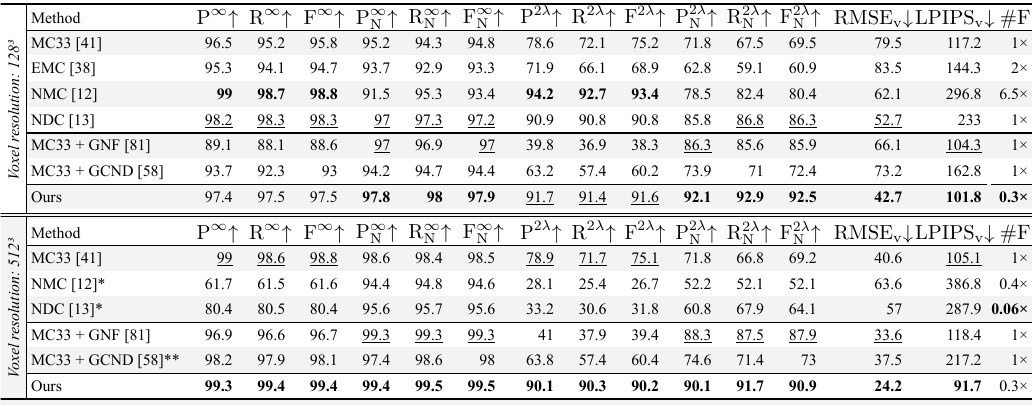}
\caption{Quantitative evaluation of mesh extraction and re-meshing performance for datasets of meshes extracted from exact SDF sampled over $128^3$ and $512^3$ volumetric grids. 
Distance- and normals-based Precision, Recall, and F-score are in percentages where higher values correspond to better results.
Measures marked with the superscript \enquote{$^{2\lambda}$} are computed over areas within $2\lambda$ to features. 
A threshold $\lambda / 16$ is used for computing distance-based measures, and a threshold of $10\degree$ for computing normals-based measures. 
$\text{RMSE}_{\text{v}}$ and $\text{LPIPS}_{\text{v}}$ are in thousandths; their lower values mean better results.
The best result is in bold, the second best is underlined.
}
\begin{tablenotes}
\scriptsize
\item[*] * For NMC~\cite{chen2021neural} and NDC~\cite{chen2022neural}, output meshes have been computed at their highest available resolution, $128^3$; 
\item[**] ** GCND~\cite{shen2022gcn} produced corrupted output for 93 out of 167 meshes, we give results for the 74 valid meshes.
\end{tablenotes}
\label{tab:sdf-extraction-statistics}
\end{table*}

\noindent \textbf{Snapping Vertices to Features.}
Assuming that smooth surface regions are well approximated by the mesh extraction method (\eg, MC33~\cite{lorensen1987marching,lewiner2003efficient}), we focus on optimizing per-point and normals deviations in regions proximal to features. 
For per-point deviations captured by $\widehat{d}^{\varepsilon}$, we identify vertices predicted to be near sharp features (vertices $v$ with $\widehat{d}^{\varepsilon}(v) \leqslant \alpha_{\text{prox}} = 2\samplingdistance $) and move them according to the \emph{vertex snapping} rule: $v \leftarrow v + \widehat{d}^{\varepsilon}(v) \cdot \widehat{r}^{\varepsilon}(v)$. 
This vertex update can be understood as a gradient descent step for minimizing $\sum_v \widehat{d}^{\varepsilon}(v)$, with $\widehat{r}^{\varepsilon}$ playing the role of the negative gradient and $\widehat{d}^{\varepsilon}$ the role of the magnitude. 
Some vertex updates violate invariants and are discarded in accordance with the general framework.
While this procedure already reconstructs features and reduces per-point deviations, the resulting mesh may still exhibit irregular triangle orientations. 
To address this, we perform the following two steps.

\noindent \textbf{Patch-Based Surface Improvement Estimation and Fusion.}
We reconstruct the surface improvement field~$\widehat{s}$ at edges in the complete mesh by predicting local fields~$\widehat{s}_i$ in each snapped patch~$P_i$ using our surface improvement net~$f_{\text{s}}$ and averaging predictions across overlapping patches.

\noindent \textbf{Interleaved Remeshing and Re-evaluation.}
To optimize surface orientation as evaluated by $\widehat{s}$, we keep the vertices in the coarse mesh fixed. We then select a subset of $n_{\text{flip}}$ edges in $E$ that are (1) predicted to yield maximal surface improvement and (2) exhibit no influence on one another (i.e., flipping an edge $e_i$ from the set should not affect the estimates $\widehat{s}(e_j)$ if $i \neq j$). Then we perform all allowed flips.
These steps are repeated until all edges with a non-negligible surface improvement value (0.01) are flipped, or a maximum number of iterations is reached. 
Following this, we re-estimate an updated surface improvement across the complete coarse mesh. 
This process can be seen as a simplified version of discrete, combinatorial minimization of an energy function of the form $\sum_{e \in E} \widehat{s}(e)$.

\noindent \textbf{Postprocessing. }
We finally perform a simple but effective postprocessing step.
For triangles oriented differently to all three of their edge-sharing neighbors, we try flipping each edge to find if this improves alignment \wrt neighbors; if so, we keep the flip in the final result.

\begin{figure}[hb!]
\centerline{\includegraphics[
width=0.9\columnwidth,]{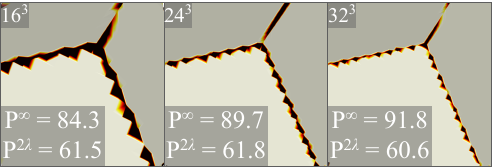}}
\caption{
Increasing volumetric resolution reduces but not eliminates aliasing. 
Global measures (Precision $\mathrm{P}^{\infty}$) are strongly influenced by resolution and therefore unsuitable for capturing feature remeshing performance. 
In contrast, our measures (Precision $\mathrm{P}^{\delta}$) focus on features and only subtly changes with resolution for aliased meshes.
}
\label{fig:resolution_dependent_measures}
\end{figure}

\begin{figure*}[htb!]
\centerline{\includegraphics[width=\textwidth]{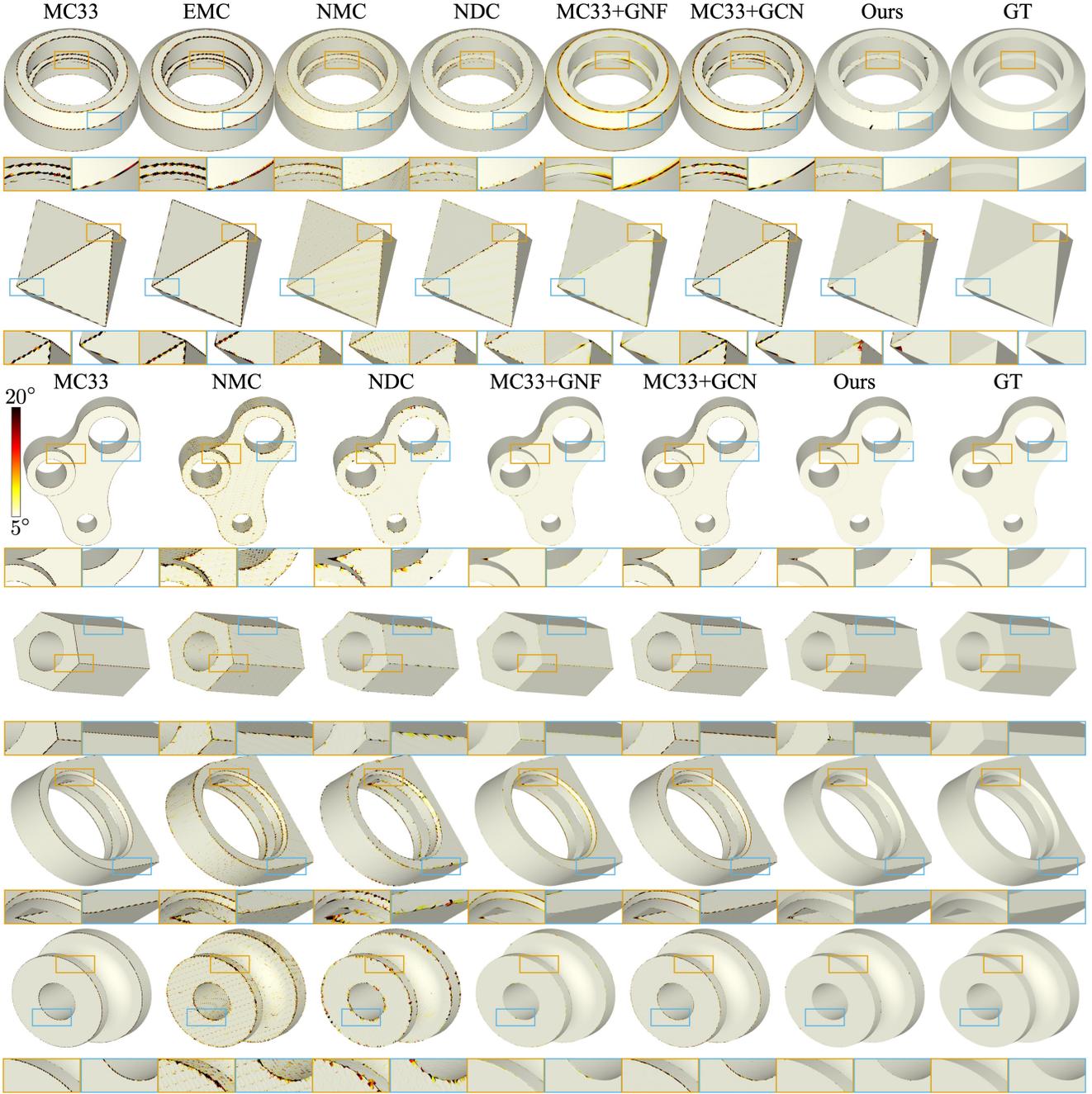}}
\caption{Qualitative reconstruction and re-meshing results for meshes extracted from SDFs sampled over $128^3$ (top two rows) and $512^3$ (bottom five rows) volumetric grids. 
We color predictions by pointwise normals difference between samples on reconstructions and their closest points on GT meshes, clipping these values to the range $[5\degree, 20\degree]$.
Lighter colors correspond to better results. 
Best viewed in zoom. 
}
\label{fig:mesh-extraction-gallery}
\end{figure*}

\section{Experimental Evaluation}
\label{sec:experiments}

\noindent \textbf{Data and Training.}
We train and evaluate our method using high-quality, feature-aligned shapes in the ABC dataset~\cite{koch2019abc}, following \cref{sec:datasets} to compute SDF values and extract coarse meshes. 
For training, we sample 36K training, 9K validation, and 5K testing patches, from 761~unique shapes. 
For evaluation, we create two object-level datasets of 159 meshes extracted from $128^3$ volumetric SDF and 167 meshes extracted from high-resolution $512^3$ volumetric SDF.
Distance-/direction-to-feature networks are optimized on patches with 500 and 2K vertices using 16 V100 GPUs; the surface improvement network is trained on patches with 500 vertices using 4 V100 GPUs; all three models were trained for 2-3 days, using early stopping. 
We use Adam~\cite{kingma2014adam} with a learning rate $10^{-3}$ and exponential decay.

\noindent \textbf{Metrics.}
To capture reconstruction accuracy and completeness globally, we respectively use Precision~($\mathrm{P}$) and Recall~($\mathrm{R}$) as commonly done for evaluating 3D reconstruction~\cite{knapitsch2017tanks,voynov2023multi}; to quantify per-face orientation similarity between meshes, we also compute Normals Precision ($\mathrm{P}_{\mathrm{N}}$) and Recall ($\mathrm{R}_{\mathrm{N}}$). 
These measures may become less informative at high voxel resolutions (\cf~\cref{fig:resolution_dependent_measures}); we thus present values globally over the entire mesh, and separately over regions within a distance $\delta \propto \samplingdistance $ to GT sharp geometric features. 
We report F-scores (F, harmonic means of the pairs of respective values) as aggregate quantities.
These measures are superscripted by either $\infty$ or $\delta$ (with $\delta = 2 \samplingdistance $).
For assessing the perceptual quality of reconstructions in comparison to GT meshes, we generate shaded renders from 32 viewing directions under 6 distinct illuminations and compute perceptual \rmsev~\cite{voynov2019perceptual} and \lpipsv~\cite{zhang2018unreasonable} measures.

We assess feature estimation performance by computing $\recall(1 \samplingdistance)$ and $\fpr(1 \samplingdistance )$ (False Positives Rate) between true and estimated distance-to-feature values. 
Similarly, we calculate $\recall(10\degree)$ and $\fpr(10\degree)$ between true and estimated per-vertex direction-to-feature values. 
To better understand their impact on vertex snapping performance, we conduct remeshing and compute mesh-based measures.

\noindent \textbf{Baselines.}
We compare our method to four mesh extraction algorithms. MC33~\cite{lewiner2003efficient} is a standard trilinear interpolation-based, local iso-surfacing method. 
EMC~\cite{kobbelt2001feature} is a feature-sensitive iso-surfacing algorithm relying on edge detection through angle thresholding, with a threshold set at~$30\degree$. 
NMC~\cite{chen2021neural} and NDC~\cite{chen2022neural} are state-of-the-art learnable methods for mesh extraction from volume data, implementing trained versions of marching cubes and dual contouring, respectively, executed at the highest resolution they have been trained on ($128^3$).
In addition, we compare our method to combinations of MC33 with two feature-preserving mesh denoising approaches. 
GNF~\cite{zhang2015guided} is a representative normal filtering-based mesh smoothing method, and GCN-Denoiser (GCND)~\cite{shen2022gcn} is a state-of-the-art two-stage denoising method integrating GCNs for normal and vertex updating. 
We run both methods with default parameter settings.

\noindent \textbf{Can we improve mesh extraction results?}
We start by assessing the performance of our framework against mesh extraction baselines, comparing meshes produced by these methods with our remeshing results. 
Quantitative results are provided in \cref{tab:sdf-extraction-statistics}, and visual examples are displayed in \cref{fig:mesh-extraction-gallery}.
For shapes extracted at $128^3$ resolution, our method outperforms all baselines across all normals-based and perceptual performance measures ($\mathrm{F}_{\mathrm{N}}^{\infty}$, $\mathrm{F}_{\mathrm{N}}^{2\samplingdistance }$, \rmsev, and \lpipsv).
In particular, improvement over the best mesh extraction method NDC constitutes 6.2\% in terms of $\mathrm{F}_{\mathrm{N}}^{\infty}$ and a relative 19\% in terms of the perceptual \rmsev.
Note that our method uses 3$\times$ fewer triangles on average. 

For $512^3$, our method outperforms all methods across all measures.
Specifically, our method improves upon NDC by 19\% in terms of $\mathrm{F}^{\infty}$ and by 58.4\% near features in terms of $\mathrm{F}^{2\samplingdistance }$. 
Postprocessing MC33 extractions by GNF or GCN-D produces smoother surfaces (as reflected in $\mathrm{F}_{\mathrm{N}}^{2\samplingdistance }$) but sacrifices fidelity to the input mesh, trading pointwise for orientation accuracy. 
Our method outperforms both GNF and GCND on normals and perceptual measures (\eg, by a relative 22.6\% and 57.8\% \lpipsv, respectively).

\begin{figure}[tb!]
\centerline{\includegraphics[width=.95\columnwidth]{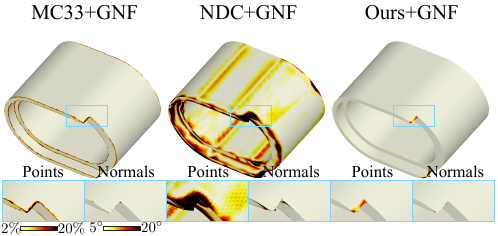}}
\caption{Qualitative mesh smoothing results. Color corresponds to pointwise distances from reconstruction to GT mesh.}
\label{fig:mesh-smoothing-gallery}
\end{figure}

\begin{table}[t]
\centering
\includegraphics[width=\columnwidth]{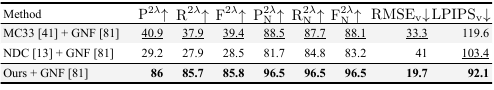}
\caption{Quantitative evaluation of mesh smoothing performance for a dataset of meshes extracted from $512^3$ volumetric SDF. 
}
\label{tab:mesh-smoothing-statistics}
\end{table}

\noindent \textbf{Can our algorithm improve mesh smoothing results?}
We study whether reconstructing more accurate geometric features by our refinement step contributes to the performance of subsequent vertex-normal filtering. 
We apply GNF to meshes obtained by the standard MC33, the state-of-the-art NDC, and the standard MC33 refined by our algorithm. 
We present statistical results for a dataset of 162 high-resolution ($512^3$) meshes in \cref{tab:mesh-smoothing-statistics}. 
We find that while mesh filtering enhances results on MC33 and NDC in terms of normals-based and perceptual measures, it struggles to fully recover sharp features due to severe aliasing in both baselines.
Our basic method (no GNF filtering) outperforms MC33, NDC, and MC33 + GNF in terms of $\mathrm{F}^{2\samplingdistance }_{\mathrm{N}}$ (\cref{tab:sdf-extraction-statistics}).
\cref{tab:mesh-smoothing-statistics} and \cref{fig:mesh-smoothing-gallery} demonstrate that adding smoothing improves our results further.

\begin{figure}[htb!]
\centerline{\includegraphics[width=.95\columnwidth]{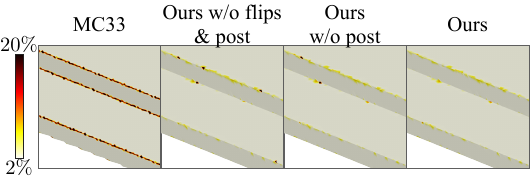}}
\caption{Qualitative ablation results. We color predictions by pointwise distances from reconstruction to GT mesh.}
\label{fig:ablation_remeshing}
\end{figure}

\begin{table}[t]
\centering
\includegraphics[width=\columnwidth]{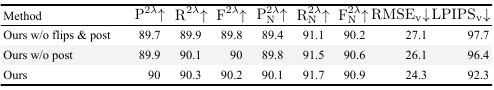}
\caption{Quantitative evaluation of contributions of the stages in our algorithm to the final performance for a dataset of meshes extracted from $512^3$ volumetric SDF. 
}
\label{tab:ablation-statistics}
\end{table}

\noindent \textbf{Which ingredients are essential for achieving good performance?}
Critical design choices for our method include (a) the data generation strategy, (b) the learning architecture, and (c) the remeshing strategy. 
Data-wise, larger patch sizes (2K vs. 500 vertices), denser voxel grids, and aggressive edge collapses prove critical, see \cref{tab:ablation-dists-dirs}. 
Learning-wise, remeshing performance is improved if the logarithmic scale is used for distances and Laplacian smoothing (\cref{sec:learning}); edge-based MeshCNN compares favorably to DGCNN adapted to operate on edges. 
For remeshing, vertex snapping, edge-flipping, and postprocessing all contribute to performance: adding flips and postprocessing increases \lpipsv by a relative 5.5\%, see~\cref{tab:ablation-statistics} and the visual result in \cref{fig:ablation_remeshing}.
We perform 2~\textit{sets of flips} in our flipping stage and display the influence of further sets in \cref{fig:quality-edge-flipping-iterations}.

\begin{figure}[htb!]
\centerline{\includegraphics[width=.7\columnwidth]{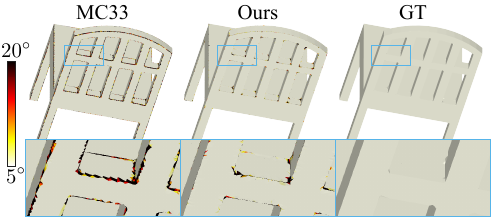}}
\caption{Generalization to ShapeNet~\cite{chang2015shapenet}. Color corresponds to normals difference between points reconstruction and GT mesh.}
\label{fig:generalization_shapenet}
\end{figure}

We additionally demonstrate a remeshed result for a high-resolution but aliased mesh from ShapeNet~\cite{chang2015shapenet} in \cref{fig:generalization_shapenet}.

\section{Conclusion}
\label{sec:conclusion}

We described a learning-based approach to feature reconstruction and remeshing that operates directly on surface 3D meshes, automatically detects sharp features in the mesh, and scales to arbitrary resolution meshes. 
Our method trains local patch-based neural models of feature-aware fields defined on the mesh and integrates these into a local remeshing approach as surface-based priors. 
Compared to state-of-the-art mesh extraction and feature-preserving mesh denoising methods, our method significantly improves surface extraction quality according to normals-based and perceptual quality measures, particularly near features. 

\section{Acknowledgements}
\label{sec:acks}

The authors acknowledge the use of Skoltech supercomputer Zhores~\cite{zacharov2019zhores} for obtaining the results presented in this paper.
The work was supported by the Analytical center under the RF Government (subsidy agreement 000000D730321P5Q0002, Grant No. 70-2021-00145 02.11.2021).

\begin{appendix}
\section*{Appendix}

\section{Details on Data Construction}
\label{sup:datasets}

\noindent \textbf{Adaptive Sampling Distance.}
To accurately represent fine-grained features in high-resolution 3D shapes (\eg, thin walls), we adapt the resolution of the volumetric grid to the complexity of the reference (ground-truth, GT) mesh.
To this end, we start with establishing a measure of complexity for a 3D shape $S$ by capturing it with a characteristic linear size~$l_S$. 
To compute $l_S$, we construct a set of all sharp geometric feature curves $\Gamma_S = \{\gamma \in S\}$ using the GT annotations available in the ABC dataset~\cite{koch2019abc}, extract their linear \textit{extents} 
\[
l_{\gamma} = \max_{n \in \{1,2,3\}} \{\max_{p \in \gamma} p_n - \min_{q \in \gamma} q_n\},
\]
(\ie, for a feature curve inscribed in a $3 \times 4 \times 5$ bounding-box, 5 will be returned),
and compute $l_S$ via as an $\alpha_{\text{curve}}$-quantile of all values $\{l_{\gamma}, \gamma \in \Gamma\}$. 
The value $l_S$ thus provides an estimate of the linear size of smaller (but not the smallest) geometric details in the shape~$S$.
As we aim to query SDF values to reference mesh in grid knots spaced~$\lambda$ units apart, and have $n$ samples
where $n$ is the number of points sampled along a curve, and $\lambda$ is a sampling distance (both are our method's parameters), we scale each mesh using the scaling factor $\beta_S$ determined by the relation $\lambda n = \beta_S l_S$.

\noindent \textbf{Mesh Simplification Details. }
Commonly, meshes extracted using iso-surfacing algorithms~\cite{lewiner2003efficient,kobbelt2001feature} include large number of small-area triangles (in particular, \enquote{needles} and \enquote{caps} with at least one angle close to zero or $\pi$).
These can be often pruned by either face or edge collapses~\cite{hoppe1996progressive}. 
In our method, we use an implementation of progressive meshes~\cite{hoppe1996progressive} available in~\cite{jiang2022declarative} and remove edges to until 33\% of input faces remain. 
Table~1 in the main text lists values 0, 0.5, and 1 for \enquote{Collapse} parameter; this refers to collapsing either no edges, half of the shortest edges (leaving 66\% of the input faces), or all shortest edges (leaving 33\% of the input faces).
The last option proves to be the optimal one according to the ablative studies (Table~1).

\noindent \textbf{Details on Computing Surface Improvement Field. }
We compute surface improvement field by approximating the integral in Equation~2 using sampling.
Specifically, we densely and uniformly sample both the coarse and the GT meshes, query the closest points on the counterpart mesh, and compute the normals difference. 
More formally, let $P_f = \{p_i\}_{i = 1}^{n_f}$ be the set of points sampled on the face $f$ in the coarse mesh. 
We compute
\begin{equation}
\label{eq:face_normals_consistency}
\mathrm{nc}(f) = \frac{1}{n_f} \sum_{i = 1}^{n_f}
    \left\lVert
    n_{\text{crs}}(p_i) - 
      n_{\text{gt}}(q_i) 
    \right\rVert 
\end{equation}
where $q_i = \pi_{\text{c}\to\text{g}}(p_i)$ is the closest point to $p_i$ on the GT surface mesh.
To compute edge-based NC measure by weighted averaging $\mathrm{nc}(e) = w_1 \mathrm{nc}(f_1) + w_2 \mathrm{nc}(f_2)$, we use weights $w_i = n_i / (n_1 + n_2), i=1, 2$ corresponding to the fraction of points sampled on $f_i, i=1,2$.

\noindent \textbf{Parameter Specifications. }
We use the following parameters to construct coarse meshes: 
\begin{itemize}
\item Sampling distance $\lambda = 0.05$ (could be arbitrarily chosen),
\item Number of samples per characteristic linear size, $n = 80$,
\item The quantile charactirizing short geometric feature curves, $\alpha_{\text{curve}} = 25\%$ ,
\item Quantile for shortest edge collapses $\alpha_{\text{short}}$ is selected to leave 33\% of input faces remain,
\item Mean inter-seed distance as supplied to Poisson Disk Sampling~\cite{bowers2010parallel} for seeds during patch cropping, $\alpha_{\text{patch}} \cdot \lambda$, $\alpha_{\text{seed}} = 16$,
\item Patch radius, $\alpha_{\text{patch}} \cdot \lambda$, $\alpha_{\text{seed}} = 32$,
\end{itemize}

\noindent \textbf{Refinement-Based Data Augmentation} is necessary as local mesh editing is inherently an iterative process leading to a gradual distribution shift for local surface patterns. 
For accurate remeshing, we require our models to operate efficiently on both sets of raw input and partically refined meshes. 
To this end, we augment our training dataset with meshes resulting from applying a series of vertex snapping and edge flipping operations computed on both ground-truth and predicted guidance fields (Section~4 in the main text).

\section{Details on Learning Configurations}
\label{sup:learning}

\noindent \textbf{Learning Architectures for Field Estimators.}
We define our learning architecture for distance- and direction-to-feature fields, inspired by DGCNN~\cite{wang2019dynamic}, as a series of parameterized point processing blocks comprised of neighbour queries, local transform, stacked point convolutions, aggregation, and interpolation layers. 
We have investigated the influence of various factors on the performance of our distance-/direction-to-feature estimators, most importantly, network capacity (depth expressed as number of layer blocks and width expressed as number of convolutional filters), its receptive field (captured by the number~$k$ of nearest neighbours), supervision regime (predicting the two feature-aware fields jointly or independently), and having different loss functions. 
We report the main findings in Table~1 in the main text and supplement them by results concerning the performance of various architectural options in \cref{tab:dgcnn-architecture}.
The best-performing distance-to-feature estimator is trained using a combination of the $L_2$ loss and the Laplacian smoothing loss, using the supervision from GT distances only, on datasets of size at least 32K patches. 

We train our direction-to-feature model using the loss
\[
L(d^{\varepsilon}, \widehat{d}^{\varepsilon}) = ||d^{\varepsilon} - \widehat{d}^{\varepsilon}||^2.
\]

\noindent \textbf{Learning Configurations for Surface Improvement Estimators.}
We train our models for estimating the edge-based surface improvement field using the surface meshes resulting from snapping vertices to sharp geometric features.

Our basic comparison is between the MeshCNN~\cite{hanocka2019meshcnn} and the DGCNN~\cite{wang2019dynamic} neural architectures. 
To adapt DGCNN to work on edges, we combine representations of two vertices sharing an edge into the edge embedding by averaging.
We additionally compare two variants of MeshCNN~\cite{hanocka2019meshcnn} architecture differing by how much in-network pooling is performed, influencing the efficiency of in-network information flow.
The default architectural option pools 2280 input edges down to 600 edges; our final choice for number of edges (1200 edges) is limited by the fact that we are training on patches where pooling boundary edges can lead to undesired behavior for the MeshCNN network. 
We experimented with the amount of pooling and found that pooling down to 1200 edges performs better than a larger variant (1600 edges) while affecting only interior edges for patches with 2280 input edges.
We present the summary of our findings in \cref{tab:ablation-flips} and discuss them in~\cref{sup:ablations}. 

Models for the improvement predictions are trained on the dataset of patches with snapped vertices. 
We explore two variants for performing vertex snaps, specifically, by using the ground-truth distance- and direction-to-feature fields, and using the predicted fields. 
After snapping vertices, we compute GT edge-based improvement values by flipping edges and calculating differences in edge-based normals consistency before and after each flip.

To account for class imbalance during training, we perform experiments using weighted binary cross-entropy (WCE)~\cite{sudre2017generalised} in addition to the default binary cross-entropy (BCE), defining WCE by 
\begin{equation}
\label{eq:wce_for_flips}
\textrm{WCE} = - \frac{1}{N}\sum_{e=1}^N ws_e\log{\widehat{s}_e} + (1 - s_e)\log(1 - \widehat{s}_e)
\end{equation}
where $w = (N - \sum_n \widehat{s}_e) / \sum_n \hat{s}_e$. 

In summary the best configuration was a MeshCNN model with pooling size of 1200, trained on patches with vertices snapped according to the values of GT distance- and direction-to-feature fields, using plain BCE loss.

\begin{table*}[t]
\centering
\includegraphics[width=\textwidth]{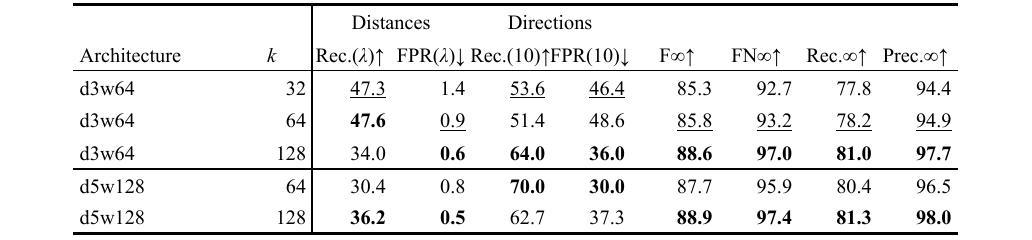}
\caption{Quantitative comparison of learning architectures for distance- and directions-to-feature approximation. 
We report edge-segmentation intersection-over-union (IoU) measure as it faithfully reflects performance for unbalanced two-class classification of surface improvement values. 
Architecture: we investigate using variants of DGCNN~\cite{wang2019dynamic} architectures with various depth and width,
$k$: number of of nearest neighbors related to the receptive field of the model.
}
\label{tab:dgcnn-architecture}
\end{table*}

\section{Details on Mesh Refinement}
\label{sup:remeshing}

\noindent \textbf{Postprocessing Details.}
We note that despite our surface improvement field is able to effectively guide the execution of a few (2--4) sets of flips (in each set, as many as possible non-interacting flips are applied), after applying these, the surface may still benefit from further remeshing. 
However, once the initial sets of flips are applied, predicting further flips becomes an extremely challenging task as the remaining target flips are too few, while meshes (mesh patches) are becoming partially refined. 
In machine learning terms, we're facing a grossly unbalanced segmentation task defined on a sampling distribution that has substantial discrepancy \wrt the data distribution of training data for predicting edge-flips.

We therefore design a simple, effective non-learnable postprocesing step that relies on the assumption that most vertices are already snapped and most edge-flips already applied. 
Specifically, we assume that most meshes we consider are formed of smooth regions connected by sharp geometric feature curves, hence, each triangle should have at least one adjacent (\ie, sharing an edge) neighboring triangle with nearly identical normal. 
If a particular triangle has no such neighbors, we conclude this to be a result of an irregular chamfering and make at attempt to flip one of its edges to improve alignment in normals to the adjacent triangles.

Concretely, we find triangles where the maximum dot-product between normals in the triangle and its adjacent neighbors is below .95, attempt to flip each of its edges, and keep flips that improve the sum of such dot-products computed over all three sides.
We perform this operation until no such triangles are left or until all flips become prohibited by invariants.

\section{Experimental Evaluation Details}
\label{sup:visuals}

\noindent \textbf{Distance-to-Feature Prediction Evaluation. }
We specify in detail additional performance measures used to assess the effectiveness of feature estimation. 
In our evaluation, we re-use the performance measures defined by the recent feature estimation method~\cite{matveev2022def}.
\begin{itemize}
\item \textit{$\recall(T)$}: we compute Recall using the thresholded predictions $\widehat{y}_i = \mathds{1}(\widehat{d}_i < T)$ and the thresholded GT distances $y_i = \mathds{1}(d_i < T)$ (we set $T = \samplingdistance$, where $\samplingdistance$ is the sampling distance, i.e. the linear voxel size).
Recall evaluates the percentage of vertices that are both located within the distance $T$ to the GT feature line and are predicted to be within $T$ to the feature line.

\item \textit{$\fpr(T)$}: we compute the False Positives Rate using the thresholded predictions and report mean \fpr across patches or full models.
\fpr estimates the percentage of vertices predicted to be located within $T$ to a sharp feature but are actually located outside the distance $T$ of the GT feature.

\item \textit{\rmse}: the root mean squared error between the predicted distances $\widehat{d}(v)$ and the GT distances $d(v)$. 

\item \textit{\qrmse}: the 95\% quantile value of \rmse across a set of instances.
\end{itemize}
We report the mean value of each measure computed across test patches.

\noindent \textbf{Direction-to-Feature Prediction Evaluation. }
We use the following measures to assess performance of direction-to-feature estimation: 
\begin{itemize}
\item \textit{$\recall(T)$}: we use the predicted direction-to-feature values $\widehat{r}(v)$ and their GT counterparts $r(v)$, and compute \recall as the percentage of vertices where angular difference between the two values does not exceed $T$: $\recall(T) = |\{v: \mathrm{arccos}(\widehat{r}(v), r(v)) < T \}| / |V|$.

\item \textit{$\fpr(T)$}: for directions, we set $\fpr(T) = 1 - \recall(T)$.

\item \textit{\rmse}: the root mean squared error between the predicted directions $\widehat{r}(v)$ and the GT directions $r(v)$. 

\item \textit{\qrmse}: the 95\% quantile value of \rmse across a set of instances.
\end{itemize}
We report the mean value of each measure computed across test patches.

\noindent \textbf{Computing Mesh-based Performance Measures. }
All of the mesh-based performance measures that we use are computed by following the same five-step process. 
Specifically, we 
\begin{enumerate}
\item[(1)] densely and uniformly sample either the reconstruction or the GT mesh, 
\item[(2)] find the closest points on the counterpart mesh, 
\item[(3)] for computing resolution-indepentent measures over regions within the distance $\delta$ to GT sharp geometric features, we exclude points that are either sampled on the GT mesh outside the distance $\delta$ to the GT features, or are sampled on the coarse mesh but have their closest points on the GT mesh outside the distance $\delta$ to the GT features,
\item[(4)] compute Euclidean distances either in 3D or normals space, 
\item[(5)] compute percentage of points with respective distances to the counterpart lower than a threshold. 
\end{enumerate}

To evaluate our feature estimation method using mesh-based performance measures, we predict (depending on the configuration) either only distances, only directions, or both distances and directions using the trained network, complement the predicted values by GT quantities where applicable (\eg, if only distances are predicted, then their predictions are supplemented by GT directions), and perform vertex snaps. 
We then sample points and compute mesh-based metrics as described above, without performing other remeshing operations.

\noindent \textbf{Implementation Details. }
We use Python's scientific computing frameworks~\cite{harris2020array,virtanen2020scipy,pedregosa2011scikit} to construct implementations for the data generation, network training, and the remeshing algorithm. 
We use PyTorch~\cite{paszke2019pytorch} to construct and optimize our deep learning models for sharp feature priors and take advantage of its higher-level neural network API PyTorch Lightning~\cite{falcon2019pytorch} to rapidly implement experimental options. 
For configuring all learning and remeshing experiments, we use the Hydra framework~\cite{Yadan2019Hydra}. 
To make our framework and comparisons reproducible, we containerize all our code in Docker~\cite{merkel2014docker}.
For mesh simplification, we use wild-meshing~\cite{jiang2022declarative}. 
We also use Point Cloud Utils~\cite{point-cloud-utils} and Trimesh~\cite{trimesh} to implement various operations involving manipulations with triangle meshes and point sampling.

\section{Additional Qualitative Results}
\label{sup:visuals}

\begin{figure*}[htb!]
\centerline{\includegraphics[width=\textwidth]{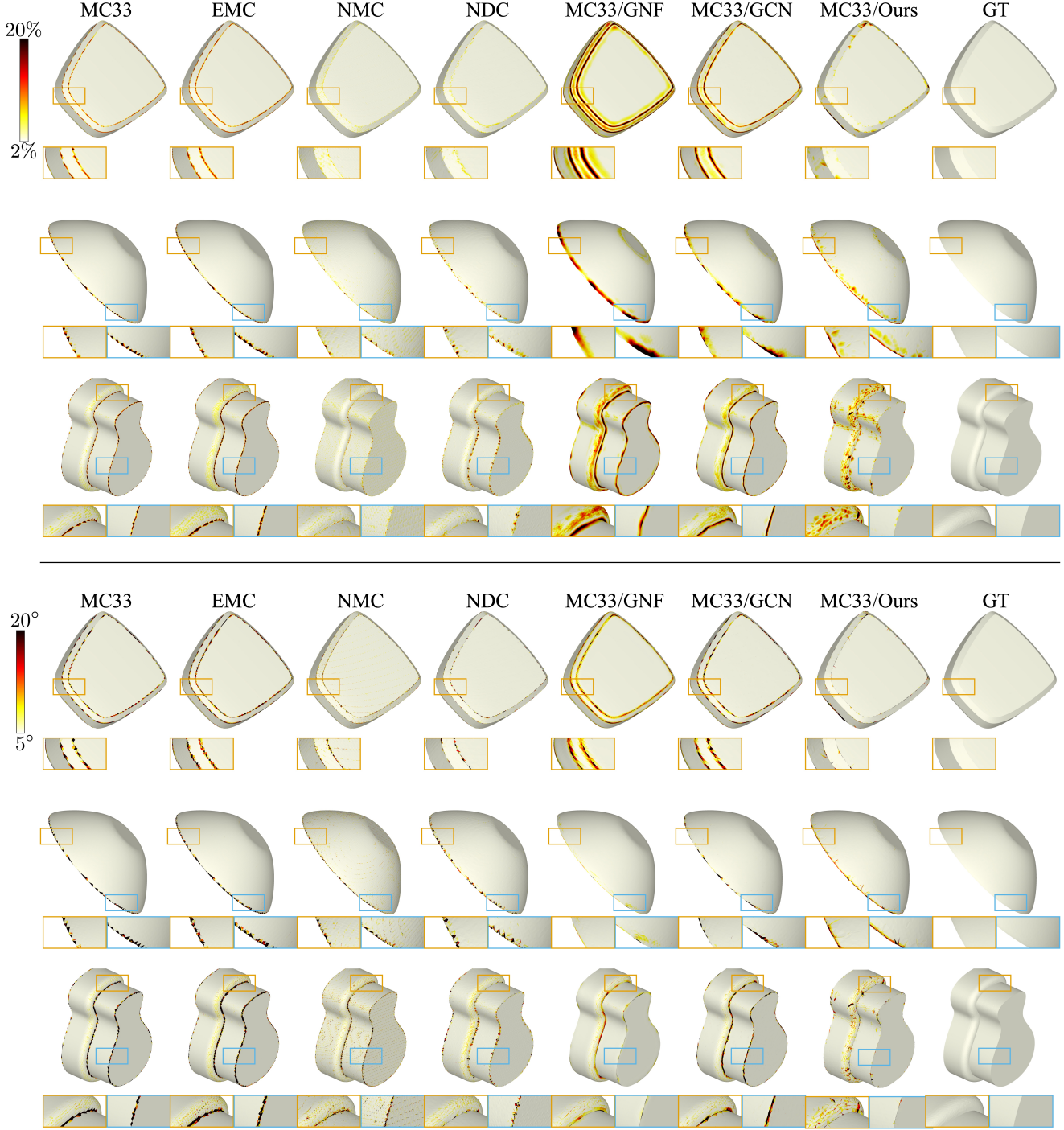}}
\caption{Qualitative reconstruction and re-meshing results for meshes extracted from SDFs sampled over $128^3$ volumetric grids. 
In \textit{top three rows,} we color predictions by pointwise distances between samples on reconstructions and their closest points on GT meshes, clipping these values to the range $[2\%, 20\%]$ relative to $\samplingdistance$.
In \textit{bottom three rows,} we color predictions by deviation in angle between normals for samples on reconstructions and their closest points on GT meshes, clipping these values to the range $[5\degree, 20\degree]$.
Lighter colors correspond to better results. 
Best viewed in zoom. 
}
\label{supp:fig-mesh-extraction-gallery-128}
\end{figure*}

\begin{figure*}[htb!]
\centerline{\includegraphics[width=.8\textwidth]{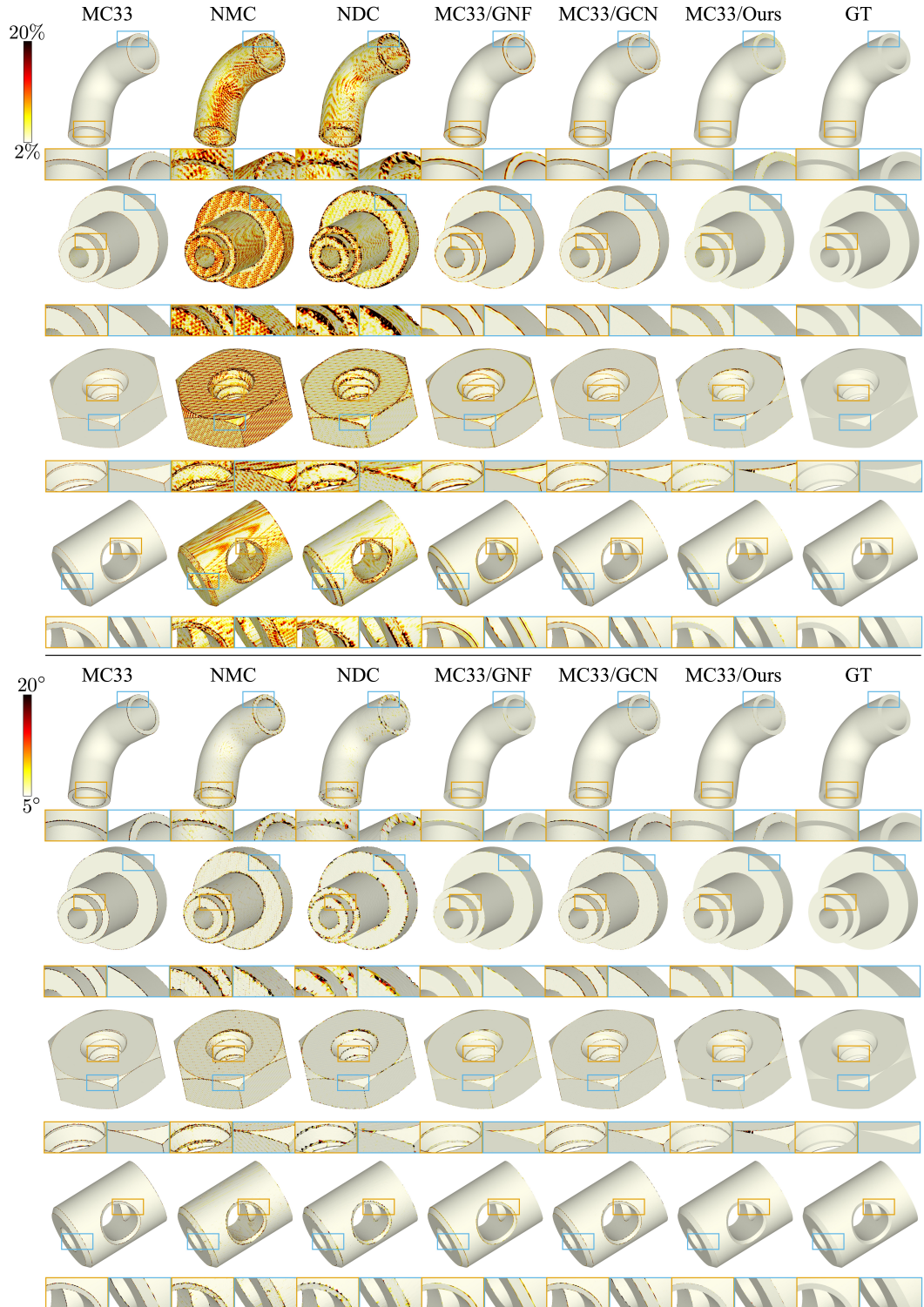}}
\caption{Qualitative reconstruction and re-meshing results for meshes extracted from SDFs sampled over $512^3$ volumetric grids. 
In \textit{top three rows,} we color predictions by pointwise distances between samples on reconstructions and their closest points on GT meshes, clipping these values to the range $[2\%, 20\%]$ relative to $\samplingdistance$.
In \textit{bottom three rows,} we color predictions by deviation in angle between normals for samples on reconstructions and their closest points on GT meshes, clipping these values to the range $[5\degree, 20\degree]$.
Lighter colors correspond to better results. 
Best viewed in zoom. 
}
\label{supp:fig-mesh-extraction-gallery-512}
\end{figure*}

We present additional qualitative results for mesh extraction and re-meshing in~\cref{supp:fig-mesh-extraction-gallery-128,supp:fig-mesh-extraction-gallery-512}.
We note that our approach consistently produces lower distortion 3D reconstructions as captured by both distance-based and normals-based performance measures.

\section{Additional Ablative Results}
\label{sup:ablations}

\begin{table}[t]
\centering
\includegraphics[width=.95\columnwidth]{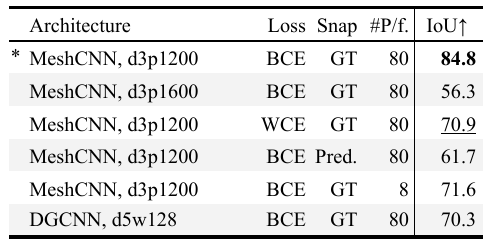}
\caption{Quantitative comparison of learning configurations for surface improvement approximation. 
We report edge-segmentation intersection-over-union (IoU) measure as it faithfully reflects performance for unbalanced two-class classification of surface improvement values. 
Architecture: we investigate using variants of MeshCNN~\cite{hanocka2019meshcnn} and DGCNN~\cite{wang2019dynamic} architectures,
Loss: binary (BCE) or weighted (WCE) cross-entropy between predictions and ground-truth segmentation masks,
Snap: for vertex snapping, we use either ground-truth (GT) or predicted (Pred.) distance- and direction-to-feature fields,
\#P/f.: points per feature.
}
\label{tab:ablation-flips}
\end{table}

\noindent \textbf{Learning Configurations for Surface Improvement. }
\cref{tab:ablation-flips} illustrates our search for an optimal configuration for surface improvement estimation.
In terms of comparison of various learning architectures, MeshCNN outperforms the edge-based DGCNN variant by 14.5\% IoU, indicating that a neural architecture specially designed to operate on edges is more suitable for edge-based estimation compared to simply \enquote{interpolating} quantities from vertices to edges.
Since we are using MeshCNN on patches and not full meshes, we were limited on the number of valid pooling operations. 
We found that pooling patches of size 2280 edges to 1200 edges was optimal, as more pooling resulted in an undefined behavior on patch boundaries, while less pooling leads to 28.5\% IoU decrease. 
Learning with a weighted loss from~\eqref{eq:wce_for_flips}, somewhat surprisingly, did not lead to advantages. 
In terms of data sources, using GT values of feature-aware fields for performing snapping resulted in 23.1\% better IoU scores, presumably due to the spurious snaps negatively impacting the robustness of flips.
As with distance- and direction-to-feature fields, larger number of samples per feature resulted in better performance.

\noindent \textbf{Learning Configurations for Feature Learning. }
In \cref{tab:dgcnn-architecture}, we present quantitative results comparing variants of our DGCNN model. 
Learning-wise, deeper architectures with larger receptive field achieve better performance; we use $k = 128$ in our final model. 
 
\end{appendix}

{\small
\bibliographystyle{ieeenat_fullname}
\bibliography{references}

\begin{thebibliography}{96}
\providecommand{\natexlab}[1]{#1}
\providecommand{\url}[1]{\texttt{#1}}
\expandafter\ifx\csname urlstyle\endcsname\relax
  \providecommand{\doi}[1]{doi: #1}\else
  \providecommand{\doi}{doi: \begingroup \urlstyle{rm}\Url}\fi

\bibitem[Alliez et~al.(2008)Alliez, Ucelli, Gotsman, and
  Attene]{alliez2008recent}
Pierre Alliez, Giuliana Ucelli, Craig Gotsman, and Marco Attene.
\newblock Recent advances in remeshing of surfaces.
\newblock \emph{Shape analysis and structuring}, pages 53--82, 2008.

\bibitem[Armando et~al.(2020)Armando, Franco, and Boyer]{armando2020mesh}
Matthieu Armando, Jean-S{\'e}bastien Franco, and Edmond Boyer.
\newblock Mesh denoising with facet graph convolutions.
\newblock \emph{IEEE Transactions on Visualization and Computer Graphics},
  28\penalty0 (8):\penalty0 2999--3012, 2020.

\bibitem[Attene et~al.(2003)Attene, Falcidieno, Rossignac, and
  Spagnuolo]{attene2003edge}
Marco Attene, Bianca Falcidieno, Jaroslaw~R Rossignac, and Michela Spagnuolo.
\newblock Edge-sharpener: recovering sharp features in triangulations of
  non-adaptively re-meshed surfaces.
\newblock Technical report, Georgia Institute of Technology, 2003.

\bibitem[Attene et~al.(2013)Attene, Campen, and Kobbelt]{attene2013polygon}
Marco Attene, Marcel Campen, and Leif Kobbelt.
\newblock Polygon mesh repairing: An application perspective.
\newblock \emph{ACM Computing Surveys (CSUR)}, 45\penalty0 (2):\penalty0 1--33,
  2013.

\bibitem[Berger et~al.(2014)Berger, Tagliasacchi, Seversky, Alliez, Levine,
  Sharf, and Silva]{berger2014state}
Matthew Berger, Andrea Tagliasacchi, Lee Seversky, Pierre Alliez, Joshua
  Levine, Andrei Sharf, and Claudio Silva.
\newblock State of the art in surface reconstruction from point clouds.
\newblock \emph{Eurographics 2014-State of the Art Reports}, 1\penalty0
  (1):\penalty0 161--185, 2014.

\bibitem[Botsch et~al.(2010)Botsch, Kobbelt, Pauly, Alliez, and
  L{\'e}vy]{botsch2010polygon}
Mario Botsch, Leif Kobbelt, Mark Pauly, Pierre Alliez, and Bruno L{\'e}vy.
\newblock \emph{Polygon mesh processing}.
\newblock CRC press, 2010.

\bibitem[Bowers et~al.(2010)Bowers, Wang, Wei, and Maletz]{bowers2010parallel}
John Bowers, Rui Wang, Li-Yi Wei, and David Maletz.
\newblock Parallel poisson disk sampling with spectrum analysis on surfaces.
\newblock \emph{ACM Transactions on Graphics (TOG)}, 29\penalty0 (6):\penalty0
  1--10, 2010.

\bibitem[Cazals and Giesen(2006)]{cazals2006delaunay}
Fr{\'e}d{\'e}ric Cazals and Joachim Giesen.
\newblock Delaunay triangulation based surface reconstruction.
\newblock In \emph{Effective computational geometry for curves and surfaces},
  pages 231--276. Springer, 2006.

\bibitem[Chabra et~al.(2020)Chabra, Lenssen, Ilg, Schmidt, Straub, Lovegrove,
  and Newcombe]{chabra2020deep}
Rohan Chabra, Jan~E Lenssen, Eddy Ilg, Tanner Schmidt, Julian Straub, Steven
  Lovegrove, and Richard Newcombe.
\newblock Deep local shapes: Learning local sdf priors for detailed 3d
  reconstruction.
\newblock In \emph{European Conference on Computer Vision}, pages 608--625.
  Springer, 2020.

\bibitem[Chang et~al.(2015)Chang, Funkhouser, Guibas, Hanrahan, Huang, Li,
  Savarese, Savva, Song, Su, et~al.]{chang2015shapenet}
Angel~X Chang, Thomas Funkhouser, Leonidas Guibas, Pat Hanrahan, Qixing Huang,
  Zimo Li, Silvio Savarese, Manolis Savva, Shuran Song, Hao Su, et~al.
\newblock Shapenet: An information-rich 3d model repository.
\newblock \emph{arXiv preprint arXiv:1512.03012}, 2015.

\bibitem[Chen et~al.(2022{\natexlab{a}})Chen, Wei, and Wang]{chen2022geometric}
Honghua Chen, Mingqiang Wei, and Jun Wang.
\newblock Geometric and learning-based mesh denoising: A comprehensive survey.
\newblock \emph{arXiv preprint arXiv:2209.00841}, 2022{\natexlab{a}}.

\bibitem[Chen and Zhang(2021)]{chen2021neural}
Zhiqin Chen and Hao Zhang.
\newblock Neural marching cubes.
\newblock \emph{ACM Transactions on Graphics (TOG)}, 40\penalty0 (6):\penalty0
  1--15, 2021.

\bibitem[Chen et~al.(2022{\natexlab{b}})Chen, Tagliasacchi, Funkhouser, and
  Zhang]{chen2022neural}
Zhiqin Chen, Andrea Tagliasacchi, Thomas Funkhouser, and Hao Zhang.
\newblock Neural dual contouring.
\newblock \emph{ACM Transactions on Graphics (TOG)}, 41\penalty0 (4):\penalty0
  1--13, 2022{\natexlab{b}}.

\bibitem[Cheng et~al.(2008)Cheng, Wang, Li, Xu, Dang, and Jin]{cheng2008survey}
Zhi-Quan Cheng, Yanzhen Wang, Bao Li, Kai Xu, Gang Dang, and Shiyao Jin.
\newblock A survey of methods for moving least squares surfaces.
\newblock In \emph{VG/PBG{@} SIGGRAPH}, pages 9--23, 2008.

\bibitem[Chernyaev(1995)]{chernyaev1995marching}
Evgeny~V Chernyaev.
\newblock Marching cubes 33: Construction of topologically correct isosurfaces.
\newblock Technical report, 1995.

\bibitem[Crane et~al.(2013)Crane, De~Goes, Desbrun, and
  Schr{\"o}der]{crane2013digital}
Keenan Crane, Fernando De~Goes, Mathieu Desbrun, and Peter Schr{\"o}der.
\newblock Digital geometry processing with discrete exterior calculus.
\newblock In \emph{ACM SIGGRAPH 2013 Courses}, pages 1--126. 2013.

\bibitem[{Dawson-Haggerty et al.}()]{trimesh}
{Dawson-Haggerty et al.}
\newblock trimesh.

\bibitem[De~Ara{\'u}jo et~al.(2015)De~Ara{\'u}jo, Lopes, Jepp, Jorge, and
  Wyvill]{de2015survey}
Bruno~Rodrigues De~Ara{\'u}jo, Daniel~S Lopes, Pauline Jepp, Joaquim~A Jorge,
  and Brian Wyvill.
\newblock A survey on implicit surface polygonization.
\newblock \emph{ACM Computing Surveys (CSUR)}, 47\penalty0 (4):\penalty0 1--39,
  2015.

\bibitem[Digne et~al.(2011)Digne, Morel, Souzani, and Lartigue]{digne2011scale}
Julie Digne, Jean-Michel Morel, Charyar-Mehdi Souzani, and Claire Lartigue.
\newblock Scale space meshing of raw data point sets.
\newblock In \emph{Computer Graphics Forum}, pages 1630--1642. Wiley Online
  Library, 2011.

\bibitem[Falcon(2019)]{falcon2019pytorch}
WA Falcon.
\newblock Pytorch lightning.
\newblock \emph{GitHub. Note:
  https://github.com/PyTorchLightning/pytorch-lightning}, 3, 2019.

\bibitem[Fan et~al.(2009)Fan, Yu, and Peng]{fan2009robust}
Hanqi Fan, Yizhou Yu, and Qunsheng Peng.
\newblock Robust feature-preserving mesh denoising based on consistent
  subneighborhoods.
\newblock \emph{IEEE Transactions on Visualization and Computer Graphics},
  16\penalty0 (2):\penalty0 312--324, 2009.

\bibitem[Feng et~al.(2023)Feng, Shen, Yuan, and Li]{feng2023deep}
Yi-Fei Feng, Li-Yong Shen, Chun-Ming Yuan, and Xin Li.
\newblock Deep shape representation with sharp feature preservation.
\newblock \emph{Computer-Aided Design}, page 103468, 2023.

\bibitem[Gu{\'e}ziec and Hummel(1995)]{gueziec1995exploiting}
Andr{\'e} Gu{\'e}ziec and Robert Hummel.
\newblock Exploiting triangulated surface extraction using tetrahedral
  decomposition.
\newblock \emph{IEEE Transactions on visualization and computer graphics},
  1\penalty0 (4):\penalty0 328--342, 1995.

\bibitem[Hanocka et~al.(2019)Hanocka, Hertz, Fish, Giryes, Fleishman, and
  Cohen-Or]{hanocka2019meshcnn}
Rana Hanocka, Amir Hertz, Noa Fish, Raja Giryes, Shachar Fleishman, and Daniel
  Cohen-Or.
\newblock Meshcnn: a network with an edge.
\newblock \emph{ACM Transactions on Graphics (TOG)}, 38\penalty0 (4):\penalty0
  1--12, 2019.

\bibitem[Harris et~al.(2020)Harris, Millman, Van Der~Walt, Gommers, Virtanen,
  Cournapeau, Wieser, Taylor, Berg, Smith, et~al.]{harris2020array}
Charles~R Harris, K~Jarrod Millman, St{\'e}fan~J Van Der~Walt, Ralf Gommers,
  Pauli Virtanen, David Cournapeau, Eric Wieser, Julian Taylor, Sebastian Berg,
  Nathaniel~J Smith, et~al.
\newblock Array programming with numpy.
\newblock \emph{Nature}, 585\penalty0 (7825):\penalty0 357--362, 2020.

\bibitem[Hattori et~al.(2022)Hattori, Yatagawa, Ohtake, and
  Suzuki]{hattori2022learning}
Shota Hattori, Tatsuya Yatagawa, Yutaka Ohtake, and Hiromasa Suzuki.
\newblock Learning self-prior for mesh denoising using dual graph convolutional
  networks.
\newblock In \emph{Computer Vision--ECCV 2022: 17th European Conference, Tel
  Aviv, Israel, October 23--27, 2022, Proceedings, Part III}, pages 363--379.
  Springer, 2022.

\bibitem[Hildebrandt et~al.(2005)Hildebrandt, Polthier, and
  Wardetzky]{hildebrandt2005smooth}
Klaus Hildebrandt, Konrad Polthier, and Max Wardetzky.
\newblock Smooth feature lines on surface meshes.
\newblock In \emph{Symposium on geometry processing}, pages 85--90, 2005.

\bibitem[Himeur et~al.(2021)Himeur, Lejemble, Pellegrini, Paulin, Barthe, and
  Mellado]{himeur2021pcednet}
Chems-Eddine Himeur, Thibault Lejemble, Thomas Pellegrini, Mathias Paulin, Loic
  Barthe, and Nicolas Mellado.
\newblock Pcednet: A lightweight neural network for fast and interactive edge
  detection in 3d point clouds.
\newblock \emph{ACM Transactions on Graphics (TOG)}, 41\penalty0 (1):\penalty0
  1--21, 2021.

\bibitem[Hoppe(1996)]{hoppe1996progressive}
Hugues Hoppe.
\newblock Progressive meshes.
\newblock In \emph{Proceedings of the 23rd annual conference on Computer
  graphics and interactive techniques}, pages 99--108, 1996.

\bibitem[Hoppe et~al.(1993)Hoppe, DeRose, Duchamp, McDonald, and
  Stuetzle]{hoppe1993mesh}
Hugues Hoppe, Tony DeRose, Tom Duchamp, John McDonald, and Werner Stuetzle.
\newblock Mesh optimization.
\newblock In \emph{Proceedings of the 20th annual conference on Computer
  graphics and interactive techniques}, pages 19--26, 1993.

\bibitem[Huang et~al.(2022)Huang, Wen, Wang, Ren, and Jia]{huang2022surface}
Zhangjin Huang, Yuxin Wen, Zihao Wang, Jinjuan Ren, and Kui Jia.
\newblock Surface reconstruction from point clouds: A survey and a benchmark.
\newblock \emph{arXiv preprint arXiv:2205.02413}, 2022.

\bibitem[Jakob et~al.(2015)Jakob, Tarini, Panozzo, Sorkine-Hornung,
  et~al.]{jakob2015instant}
Wenzel Jakob, Marco Tarini, Daniele Panozzo, Olga Sorkine-Hornung, et~al.
\newblock Instant field-aligned meshes.
\newblock \emph{ACM Trans. Graph.}, 34\penalty0 (6):\penalty0 189--1, 2015.

\bibitem[Jiang et~al.(2020)Jiang, Sud, Makadia, Huang, Nie{\ss}ner, Funkhouser,
  et~al.]{jiang2020local}
Chiyu Jiang, Avneesh Sud, Ameesh Makadia, Jingwei Huang, Matthias Nie{\ss}ner,
  Thomas Funkhouser, et~al.
\newblock Local implicit grid representations for 3d scenes.
\newblock In \emph{Proceedings of the IEEE/CVF Conference on Computer Vision
  and Pattern Recognition}, pages 6001--6010, 2020.

\bibitem[Jiang et~al.(2022)Jiang, Dai, Hu, Zhou, Dumas, Zhou, Bajwa, Zorin,
  Panozzo, and Schneider]{jiang2022declarative}
Zhongshi Jiang, Jiacheng Dai, Yixin Hu, Yunfan Zhou, Jeremie Dumas, Qingnan
  Zhou, Gurkirat~Singh Bajwa, Denis Zorin, Daniele Panozzo, and Teseo
  Schneider.
\newblock Declarative specification for unstructured mesh editing algorithms.
\newblock \emph{ACM Transactions on Graphics (TOG)}, 41\penalty0 (6):\penalty0
  1--14, 2022.

\bibitem[Ju et~al.(2002)Ju, Losasso, Schaefer, and Warren]{ju2002dual}
Tao Ju, Frank Losasso, Scott Schaefer, and Joe Warren.
\newblock Dual contouring of hermite data.
\newblock In \emph{Proceedings of the 29th annual conference on Computer
  graphics and interactive techniques}, pages 339--346, 2002.

\bibitem[Kalogerakis et~al.(2007)Kalogerakis, Simari, Nowrouzezahrai, and
  Singh]{kalogerakis2007robust}
Evangelos Kalogerakis, Patricio Simari, Derek Nowrouzezahrai, and Karan Singh.
\newblock Robust statistical estimation of curvature on discretized surfaces.
\newblock In \emph{Symposium on Geometry Processing}, pages 110--114, 2007.

\bibitem[Kazhdan and Hoppe(2013)]{kazhdan2013screened}
Michael Kazhdan and Hugues Hoppe.
\newblock Screened poisson surface reconstruction.
\newblock \emph{ACM Transactions on Graphics (ToG)}, 32\penalty0 (3):\penalty0
  1--13, 2013.

\bibitem[Khatamian and Arabnia(2016)]{khatamian2016survey}
Alireza Khatamian and Hamid~R Arabnia.
\newblock Survey on 3d surface reconstruction.
\newblock \emph{Journal of Information Processing Systems}, 12\penalty0
  (3):\penalty0 338--357, 2016.

\bibitem[Kim et~al.(2009)Kim, Choi, and Lee]{kim2009feature}
Hyun~Soo Kim, Han~Kyun Choi, and Kwan~H Lee.
\newblock Feature detection of triangular meshes based on tensor voting theory.
\newblock \emph{Computer-Aided Design}, 41\penalty0 (1):\penalty0 47--58, 2009.

\bibitem[Kingma and Ba(2014)]{kingma2014adam}
Diederik~P Kingma and Jimmy Ba.
\newblock Adam: A method for stochastic optimization.
\newblock \emph{arXiv preprint arXiv:1412.6980}, 2014.

\bibitem[Knapitsch et~al.(2017)Knapitsch, Park, Zhou, and
  Koltun]{knapitsch2017tanks}
Arno Knapitsch, Jaesik Park, Qian-Yi Zhou, and Vladlen Koltun.
\newblock Tanks and temples: Benchmarking large-scale scene reconstruction.
\newblock \emph{ACM Transactions on Graphics (ToG)}, 36\penalty0 (4):\penalty0
  1--13, 2017.

\bibitem[Kobbelt et~al.(2001)Kobbelt, Botsch, Schwanecke, and
  Seidel]{kobbelt2001feature}
Leif~P Kobbelt, Mario Botsch, Ulrich Schwanecke, and Hans-Peter Seidel.
\newblock Feature sensitive surface extraction from volume data.
\newblock In \emph{Proceedings of the 28th annual conference on Computer
  graphics and interactive techniques}, pages 57--66, 2001.

\bibitem[Koch et~al.(2019)Koch, Matveev, Jiang, Williams, Artemov, Burnaev,
  Alexa, Zorin, and Panozzo]{koch2019abc}
Sebastian Koch, Albert Matveev, Zhongshi Jiang, Francis Williams, Alexey
  Artemov, Evgeny Burnaev, Marc Alexa, Denis Zorin, and Daniele Panozzo.
\newblock Abc: A big cad model dataset for geometric deep learning.
\newblock In \emph{Proceedings of the IEEE Conference on Computer Vision and
  Pattern Recognition}, pages 9601--9611, 2019.

\bibitem[Lai et~al.(2006)Lai, Zhou, Hu, Wallner, and Pottmann]{lai2006robust}
Yu-Kun Lai, Qian-Yi Zhou, Shi-Min Hu, Johannes Wallner, and Helmut Pottmann.
\newblock Robust feature classification and editing.
\newblock \emph{IEEE Transactions on Visualization and Computer Graphics},
  13\penalty0 (1):\penalty0 34--45, 2006.

\bibitem[Lewiner et~al.(2003)Lewiner, Lopes, Vieira, and
  Tavares]{lewiner2003efficient}
Thomas Lewiner, H{\'e}lio Lopes, Ant{\^o}nio~Wilson Vieira, and Geovan Tavares.
\newblock Efficient implementation of marching cubes' cases with topological
  guarantees.
\newblock \emph{Journal of graphics tools}, 8\penalty0 (2):\penalty0 1--15,
  2003.

\bibitem[Li et~al.(2018)Li, Zhu, Fu, and Heng]{li2018non}
Xianzhi Li, Lei Zhu, Chi-Wing Fu, and Pheng-Ann Heng.
\newblock Non-local low-rank normal filtering for mesh denoising.
\newblock In \emph{Computer Graphics Forum}, pages 155--166. Wiley Online
  Library, 2018.

\bibitem[Li et~al.(2020)Li, Li, Zhu, Fu, and Heng]{li2020dnf}
Xianzhi Li, Ruihui Li, Lei Zhu, Chi-Wing Fu, and Pheng-Ann Heng.
\newblock Dnf-net: A deep normal filtering network for mesh denoising.
\newblock \emph{IEEE Transactions on Visualization and Computer Graphics},
  27\penalty0 (10):\penalty0 4060--4072, 2020.

\bibitem[Liu et~al.(2020)Liu, Zhang, and Su]{liu2020meshing}
Minghua Liu, Xiaoshuai Zhang, and Hao Su.
\newblock Meshing point clouds with predicted intrinsic-extrinsic ratio
  guidance.
\newblock In \emph{European Conference on Computer Vision}, pages 68--84.
  Springer, 2020.

\bibitem[Liu et~al.()Liu, D'Aronco, Schindler, and Wegner]{liupc2wf}
Yujia Liu, Stefano D'Aronco, Konrad Schindler, and Jan~Dirk Wegner.
\newblock Pc2wf: 3d wireframe reconstruction from raw point clouds.
\newblock In \emph{International Conference on Learning Representations}.

\bibitem[Lorensen and Cline(1987)]{lorensen1987marching}
William~E Lorensen and Harvey~E Cline.
\newblock Marching cubes: A high resolution 3d surface construction algorithm.
\newblock \emph{ACM siggraph computer graphics}, 21\penalty0 (4):\penalty0
  163--169, 1987.

\bibitem[Matveev et~al.(2022)Matveev, Rakhimov, Artemov, Bobrovskikh,
  Egiazarian, Bogomolov, Panozzo, Zorin, and Burnaev]{matveev2022def}
Albert Matveev, Ruslan Rakhimov, Alexey Artemov, Gleb Bobrovskikh, Vage
  Egiazarian, Emil Bogomolov, Daniele Panozzo, Denis Zorin, and Evgeny Burnaev.
\newblock Def: Deep estimation of sharp geometric features in 3d shapes.
\newblock \emph{ACM Transactions on Graphics (TOG)}, 41\penalty0 (4):\penalty0
  1--22, 2022.

\bibitem[Merkel et~al.(2014)]{merkel2014docker}
Dirk Merkel et~al.
\newblock Docker: lightweight linux containers for consistent development and
  deployment.
\newblock \emph{Linux j}, 239\penalty0 (2):\penalty0 2, 2014.

\bibitem[Mescheder et~al.(2019)Mescheder, Oechsle, Niemeyer, Nowozin, and
  Geiger]{mescheder2019occupancy}
Lars Mescheder, Michael Oechsle, Michael Niemeyer, Sebastian Nowozin, and
  Andreas Geiger.
\newblock Occupancy networks: Learning 3d reconstruction in function space.
\newblock In \emph{Proceedings of the IEEE/CVF conference on computer vision
  and pattern recognition}, pages 4460--4470, 2019.

\bibitem[Newman and Yi(2006)]{newman2006survey}
Timothy~S Newman and Hong Yi.
\newblock A survey of the marching cubes algorithm.
\newblock \emph{Computers \& Graphics}, 30\penalty0 (5):\penalty0 854--879,
  2006.

\bibitem[Ohtake et~al.(2004)Ohtake, Belyaev, and Seidel]{ohtake2004ridge}
Yutaka Ohtake, Alexander Belyaev, and Hans-Peter Seidel.
\newblock Ridge-valley lines on meshes via implicit surface fitting.
\newblock In \emph{ACM SIGGRAPH 2004 Papers}, pages 609--612. 2004.

\bibitem[{\"O}ztireli et~al.(2009){\"O}ztireli, Guennebaud, and
  Gross]{oztireli2009feature}
A~Cengiz {\"O}ztireli, Gael Guennebaud, and Markus Gross.
\newblock Feature preserving point set surfaces based on non-linear kernel
  regression.
\newblock In \emph{Computer graphics forum}, pages 493--501. Wiley Online
  Library, 2009.

\bibitem[Park et~al.(2019)Park, Florence, Straub, Newcombe, and
  Lovegrove]{park2019deepsdf}
Jeong~Joon Park, Peter Florence, Julian Straub, Richard Newcombe, and Steven
  Lovegrove.
\newblock Deepsdf: Learning continuous signed distance functions for shape
  representation.
\newblock In \emph{Proceedings of the IEEE/CVF conference on computer vision
  and pattern recognition}, pages 165--174, 2019.

\bibitem[Paszke et~al.(2019)Paszke, Gross, Massa, Lerer, Bradbury, Chanan,
  Killeen, Lin, Gimelshein, Antiga, et~al.]{paszke2019pytorch}
Adam Paszke, Sam Gross, Francisco Massa, Adam Lerer, James Bradbury, Gregory
  Chanan, Trevor Killeen, Zeming Lin, Natalia Gimelshein, Luca Antiga, et~al.
\newblock Pytorch: An imperative style, high-performance deep learning library.
\newblock \emph{Advances in neural information processing systems}, 32, 2019.

\bibitem[Pedregosa et~al.(2011)Pedregosa, Varoquaux, Gramfort, Michel, Thirion,
  Grisel, Blondel, Prettenhofer, Weiss, Dubourg, et~al.]{pedregosa2011scikit}
Fabian Pedregosa, Ga{\"e}l Varoquaux, Alexandre Gramfort, Vincent Michel,
  Bertrand Thirion, Olivier Grisel, Mathieu Blondel, Peter Prettenhofer, Ron
  Weiss, Vincent Dubourg, et~al.
\newblock Scikit-learn: Machine learning in python.
\newblock \emph{the Journal of machine Learning research}, 12:\penalty0
  2825--2830, 2011.

\bibitem[Pietroni et~al.(2021)Pietroni, Nuvoli, Alderighi, Cignoni, Tarini,
  et~al.]{pietroni2021reliable}
Nico Pietroni, Stefano Nuvoli, Thomas Alderighi, Paolo Cignoni, Marco Tarini,
  et~al.
\newblock Reliable feature-line driven quad-remeshing.
\newblock \emph{ACM Transactions on Graphics}, 40\penalty0 (4):\penalty0 1--17,
  2021.

\bibitem[Qu et~al.(2004)Qu, Zhang, Shao, Kaufman, and Mueller]{qu2004feature}
Huamin Qu, Nan Zhang, Ran Shao, Arie Kaufman, and Klaus Mueller.
\newblock Feature preserving distance fields.
\newblock In \emph{2004 IEEE Symposium on Volume Visualization and Graphics},
  pages 39--46. IEEE, 2004.

\bibitem[Rakotosaona et~al.(2021)Rakotosaona, Guerrero, Aigerman, Mitra, and
  Ovsjanikov]{rakotosaona2021learning}
Marie-Julie Rakotosaona, Paul Guerrero, Noam Aigerman, Niloy~J Mitra, and Maks
  Ovsjanikov.
\newblock Learning delaunay surface elements for mesh reconstruction.
\newblock In \emph{Proceedings of the IEEE/CVF Conference on Computer Vision
  and Pattern Recognition}, pages 22--31, 2021.

\bibitem[Schmelling(1995)]{schmelling1995averaging}
Michael Schmelling.
\newblock Averaging correlated data.
\newblock \emph{Physica Scripta}, 51\penalty0 (6):\penalty0 676, 1995.

\bibitem[Sharp and Ovsjanikov(2020)]{sharp2020pointtrinet}
Nicholas Sharp and Maks Ovsjanikov.
\newblock Pointtrinet: Learned triangulation of 3d point sets.
\newblock In \emph{European Conference on Computer Vision}, pages 762--778.
  Springer, 2020.

\bibitem[Shen et~al.(2022)Shen, Fu, Du, Chen, Burnaev, Zorin, Zhou, and
  Zheng]{shen2022gcn}
Yuefan Shen, Hongbo Fu, Zhongshuo Du, Xiang Chen, Evgeny Burnaev, Denis Zorin,
  Kun Zhou, and Youyi Zheng.
\newblock Gcn-denoiser: Mesh denoising with graph convolutional networks.
\newblock \emph{ACM Transactions on Graphics (TOG)}, 41\penalty0 (1):\penalty0
  1--14, 2022.

\bibitem[Shorten and Khoshgoftaar(2019)]{shorten2019survey}
Connor Shorten and Taghi~M Khoshgoftaar.
\newblock A survey on image data augmentation for deep learning.
\newblock \emph{Journal of big data}, 6\penalty0 (1):\penalty0 1--48, 2019.

\bibitem[Sudre et~al.(2017)Sudre, Li, Vercauteren, Ourselin, and
  Jorge~Cardoso]{sudre2017generalised}
Carole~H Sudre, Wenqi Li, Tom Vercauteren, Sebastien Ourselin, and M
  Jorge~Cardoso.
\newblock Generalised dice overlap as a deep learning loss function for highly
  unbalanced segmentations.
\newblock In \emph{Deep Learning in Medical Image Analysis and Multimodal
  Learning for Clinical Decision Support: Third International Workshop, DLMIA
  2017, and 7th International Workshop, ML-CDS 2017, Held in Conjunction with
  MICCAI 2017, Qu{\'e}bec City, QC, Canada, September 14, Proceedings 3}, pages
  240--248. Springer, 2017.

\bibitem[Sun et~al.(2007)Sun, Rosin, Martin, and Langbein]{sun2007fast}
Xianfang Sun, Paul~L Rosin, Ralph Martin, and Frank Langbein.
\newblock Fast and effective feature-preserving mesh denoising.
\newblock \emph{IEEE transactions on visualization and computer graphics},
  13\penalty0 (5):\penalty0 925--938, 2007.

\bibitem[Surazhsky and Gotsman(2003)]{surazhsky2003explicit}
Vitaly Surazhsky and Craig Gotsman.
\newblock Explicit surface remeshing.
\newblock In \emph{Proceedings of the 2003 Eurographics/ACM SIGGRAPH symposium
  on Geometry processing}, pages 20--30, 2003.

\bibitem[Tarini et~al.(2010)Tarini, Pietroni, Cignoni, Panozzo, and
  Puppo]{tarini2010practical}
Marco Tarini, Nico Pietroni, Paolo Cignoni, Daniele Panozzo, and Enrico Puppo.
\newblock Practical quad mesh simplification.
\newblock In \emph{Computer Graphics Forum}, pages 407--418. Wiley Online
  Library, 2010.

\bibitem[Taubin et~al.(2001)]{taubin2001linear}
Gabriel Taubin et~al.
\newblock Linear anisotropic mesh filtering.
\newblock \emph{Res. Rep. RC2213 IBM}, 1\penalty0 (4), 2001.

\bibitem[Thompson et~al.(2019)Thompson, Arvanitis, Moustakas, Hoang-Xuan,
  Nguyen, Tran, Lejemble, Barthe, Mellado, Romanengo,
  et~al.]{thompson2019shrec}
E~Moscoso Thompson, G Arvanitis, K Moustakas, N Hoang-Xuan, ER Nguyen, M Tran,
  Thibault Lejemble, Loic Barthe, Nicolas Mellado, C Romanengo, et~al.
\newblock Shrec'19 track: Feature curve extraction on triangle meshes.
\newblock In \emph{12th EG Workshop 3D Object Retrieval 2019}, pages 1--8,
  2019.

\bibitem[Torrente et~al.(2018)Torrente, Biasotti, and
  Falcidieno]{torrente2018recognition}
Maria-Laura Torrente, Silvia Biasotti, and Bianca Falcidieno.
\newblock Recognition of feature curves on 3d shapes using an algebraic
  approach to hough transforms.
\newblock \emph{Pattern Recognition}, 73:\penalty0 111--130, 2018.

\bibitem[Treece et~al.(1999)Treece, Prager, and Gee]{treece1999regularised}
Graham~M Treece, Richard~W Prager, and Andrew~H Gee.
\newblock Regularised marching tetrahedra: improved iso-surface extraction.
\newblock \emph{Computers \& Graphics}, 23\penalty0 (4):\penalty0 583--598,
  1999.

\bibitem[Varadhan et~al.(2003)Varadhan, Krishnan, Kim, and
  Manocha]{varadhan2003feature}
Gokul Varadhan, Shankar Krishnan, Young~J Kim, and Dinesh Manocha.
\newblock Feature-sensitive subdivision and isosurface reconstruction.
\newblock In \emph{IEEE Visualization, 2003. VIS 2003.}, pages 99--106. IEEE,
  2003.

\bibitem[Virtanen et~al.(2020)Virtanen, Gommers, Oliphant, Haberland, Reddy,
  Cournapeau, Burovski, Peterson, Weckesser, Bright, et~al.]{virtanen2020scipy}
Pauli Virtanen, Ralf Gommers, Travis~E Oliphant, Matt Haberland, Tyler Reddy,
  David Cournapeau, Evgeni Burovski, Pearu Peterson, Warren Weckesser, Jonathan
  Bright, et~al.
\newblock Scipy 1.0: fundamental algorithms for scientific computing in python.
\newblock \emph{Nature methods}, 17\penalty0 (3):\penalty0 261--272, 2020.

\bibitem[Vorsatz et~al.(2001)Vorsatz, R{\"o}ssl, Kobbelt, and
  Seidel]{vorsatz2001feature}
Jens Vorsatz, Christian R{\"o}ssl, Leif~P Kobbelt, and H-P Seidel.
\newblock Feature sensitive remeshing.
\newblock In \emph{Computer Graphics Forum}, pages 393--401. Wiley Online
  Library, 2001.

\bibitem[Voynov et~al.(2019)Voynov, Artemov, Egiazarian, Notchenko,
  Bobrovskikh, Burnaev, and Zorin]{voynov2019perceptual}
Oleg Voynov, Alexey Artemov, Vage Egiazarian, Alexander Notchenko, Gleb
  Bobrovskikh, Evgeny Burnaev, and Denis Zorin.
\newblock Perceptual deep depth super-resolution.
\newblock In \emph{Proceedings of the IEEE/CVF International Conference on
  Computer Vision}, pages 5653--5663, 2019.

\bibitem[Voynov et~al.(2023)Voynov, Bobrovskikh, Karpyshev, Galochkin,
  Ardelean, Bozhenko, Karmanova, Kopanev, Labutin-Rymsho, Rakhimov,
  et~al.]{voynov2023multi}
Oleg Voynov, Gleb Bobrovskikh, Pavel Karpyshev, Saveliy Galochkin,
  Andrei-Timotei Ardelean, Arseniy Bozhenko, Ekaterina Karmanova, Pavel
  Kopanev, Yaroslav Labutin-Rymsho, Ruslan Rakhimov, et~al.
\newblock Multi-sensor large-scale dataset for multi-view 3d reconstruction.
\newblock In \emph{Proceedings of the IEEE/CVF Conference on Computer Vision
  and Pattern Recognition}, pages 21392--21403, 2023.

\bibitem[Wang et~al.(2012)Wang, Zhang, and Yu]{wang2012cascaded}
Jun Wang, Xi Zhang, and Zeyun Yu.
\newblock A cascaded approach for feature-preserving surface mesh denoising.
\newblock \emph{Computer-Aided Design}, 44\penalty0 (7):\penalty0 597--610,
  2012.

\bibitem[Wang et~al.(2016)Wang, Liu, and Tong]{wang2016mesh}
Peng-Shuai Wang, Yang Liu, and Xin Tong.
\newblock Mesh denoising via cascaded normal regression.
\newblock \emph{ACM Trans. Graph.}, 35\penalty0 (6):\penalty0 232--1, 2016.

\bibitem[Wang et~al.(2020)Wang, Xu, Xu, Tagliasacchi, Zhou, Mahdavi-Amiri, and
  Zhang]{wang2020pie}
Xiaogang Wang, Yuelang Xu, Kai Xu, Andrea Tagliasacchi, Bin Zhou, Ali
  Mahdavi-Amiri, and Hao Zhang.
\newblock Pie-net: Parametric inference of point cloud edges.
\newblock \emph{Advances in neural information processing systems},
  33:\penalty0 20167--20178, 2020.

\bibitem[Wang et~al.(2019)Wang, Sun, Liu, Sarma, Bronstein, and
  Solomon]{wang2019dynamic}
Yue Wang, Yongbin Sun, Ziwei Liu, Sanjay~E Sarma, Michael~M Bronstein, and
  Justin~M Solomon.
\newblock Dynamic graph cnn for learning on point clouds.
\newblock \emph{Acm Transactions On Graphics (tog)}, 38\penalty0 (5):\penalty0
  1--12, 2019.

\bibitem[Wei et~al.(2018)Wei, Huang, Xie, Liu, Wang, and Qin]{wei2018mesh}
Mingqiang Wei, Jin Huang, Xingyu Xie, Ligang Liu, Jun Wang, and Jing Qin.
\newblock Mesh denoising guided by patch normal co-filtering via kernel
  low-rank recovery.
\newblock \emph{IEEE transactions on visualization and computer graphics},
  25\penalty0 (10):\penalty0 2910--2926, 2018.

\bibitem[Williams(2022)]{point-cloud-utils}
Francis Williams.
\newblock Point cloud utils, 2022.
\newblock https://www.github.com/fwilliams/point-cloud-utils.

\bibitem[Williams et~al.(2022)Williams, Gojcic, Khamis, Zorin, Bruna, Fidler,
  and Litany]{williams2022neural}
Francis Williams, Zan Gojcic, Sameh Khamis, Denis Zorin, Joan Bruna, Sanja
  Fidler, and Or Litany.
\newblock Neural fields as learnable kernels for 3d reconstruction.
\newblock In \emph{Proceedings of the IEEE/CVF Conference on Computer Vision
  and Pattern Recognition}, pages 18500--18510, 2022.

\bibitem[Xu et~al.(2022)Xu, Wang, Dou, Zong, Xin, Jiang, Ju, and
  Tu]{xu2022rfeps}
Rui Xu, Zixiong Wang, Zhiyang Dou, Chen Zong, Shiqing Xin, Mingyan Jiang, Tao
  Ju, and Changhe Tu.
\newblock Rfeps: Reconstructing feature-line equipped polygonal surface.
\newblock \emph{ACM Transactions on Graphics (TOG)}, 41\penalty0 (6):\penalty0
  1--15, 2022.

\bibitem[Yadan(2019)]{Yadan2019Hydra}
Omry Yadan.
\newblock Hydra - a framework for elegantly configuring complex applications.
\newblock Github, 2019.

\bibitem[Yadav et~al.(2018)Yadav, Reitebuch, and Polthier]{yadav2018robust}
Sunil~Kumar Yadav, Ulrich Reitebuch, and Konrad Polthier.
\newblock Robust and high fidelity mesh denoising.
\newblock \emph{IEEE transactions on visualization and computer graphics},
  25\penalty0 (6):\penalty0 2304--2310, 2018.

\bibitem[Yoshizawa et~al.(2008)Yoshizawa, Belyaev, Yokota, and
  Seidel]{yoshizawa2008fast}
Shin Yoshizawa, Alexander Belyaev, Hideo Yokota, and Hans-Peter Seidel.
\newblock Fast, robust, and faithful methods for detecting crest lines on
  meshes.
\newblock \emph{Computer Aided Geometric Design}, 25\penalty0 (8):\penalty0
  545--560, 2008.

\bibitem[Zacharov et~al.(2019)Zacharov, Arslanov, Gunin, Stefonishin, Bykov,
  Pavlov, Panarin, Maliutin, Rykovanov, and Fedorov]{zacharov2019zhores}
Igor Zacharov, Rinat Arslanov, Maksim Gunin, Daniil Stefonishin, Andrey Bykov,
  Sergey Pavlov, Oleg Panarin, Anton Maliutin, Sergey Rykovanov, and Maxim
  Fedorov.
\newblock “zhores”—petaflops supercomputer for data-driven modeling,
  machine learning and artificial intelligence installed in skolkovo institute
  of science and technology.
\newblock \emph{Open Engineering}, 9\penalty0 (1):\penalty0 512--520, 2019.

\bibitem[Zhang et~al.(2018)Zhang, Isola, Efros, Shechtman, and
  Wang]{zhang2018unreasonable}
Richard Zhang, Phillip Isola, Alexei~A Efros, Eli Shechtman, and Oliver Wang.
\newblock The unreasonable effectiveness of deep features as a perceptual
  metric.
\newblock In \emph{Proceedings of the IEEE conference on computer vision and
  pattern recognition}, pages 586--595, 2018.

\bibitem[Zhang et~al.(2015)Zhang, Deng, Zhang, Bouaziz, and
  Liu]{zhang2015guided}
Wangyu Zhang, Bailin Deng, Juyong Zhang, Sofien Bouaziz, and Ligang Liu.
\newblock Guided mesh normal filtering.
\newblock In \emph{Computer Graphics Forum}, pages 23--34. Wiley Online
  Library, 2015.

\bibitem[Zhang et~al.(2022)Zhang, Shen, Wang, Qian, Wei, and
  Qin]{zhang2022geobi}
Yingkui Zhang, Guibao Shen, Qiong Wang, Yinling Qian, Mingqiang Wei, and Jing
  Qin.
\newblock Geobi-gnn: geometry-aware bi-domain mesh denoising via graph neural
  networks.
\newblock \emph{Computer-Aided Design}, 144:\penalty0 103154, 2022.

\bibitem[Zhao et~al.(2019)Zhao, Liu, Zhao, Fan, and Zhao]{zhao2019normalnet}
Wenbo Zhao, Xianming Liu, Yongsen Zhao, Xiaopeng Fan, and Debin Zhao.
\newblock Normalnet: Learning based guided normal filtering for mesh denoising.
\newblock \emph{arXiv preprint arXiv:1903.04015}, 2019.

\bibitem[Zheng et~al.(2010)Zheng, Fu, Au, and Tai]{zheng2010bilateral}
Youyi Zheng, Hongbo Fu, Oscar Kin-Chung Au, and Chiew-Lan Tai.
\newblock Bilateral normal filtering for mesh denoising.
\newblock \emph{IEEE transactions on visualization and computer graphics},
  17\penalty0 (10):\penalty0 1521--1530, 2010.

\end{thebibliography}
}

\end{document}